%% file: jfrExample.tex
\pdfoutput=1
\documentclass{article}
\usepackage{jfrExamplee}
\usepackage[dvipdfmx]{graphicx}
\usepackage{apalike}
\usepackage{setspace}
\usepackage{amsmath}
\usepackage{color}
\usepackage{url}
\usepackage{dsfont}
\usepackage{comment}
\usepackage{subfig}
\usepackage{cite}

\title{Reliable Monte Carlo Localization for Mobile Robots}

\author{
Naoki Akai$^{1, 2}$\thanks{ Home page of the corresponding author: https://sites.google.com/view/naokiakaigoo/home} \\
$^{1}$Graduate School of Engineering, Nagoya University\\
Furo, Chikusa, Nagoya, Japan\\
$^{2}$LOCT Co., Ltd.\\
Ikatsu, Showa, Nagoya, Japan\\
\texttt{akai@nagoya-u.jp} \\
}

\begin{document}

\maketitle

\begin{abstract}
Reliability is a key factor for realizing safety guarantee of fully autonomous robot systems. In this paper, we focus on reliability in mobile robot localization. Monte Carlo localization (MCL) is widely used for mobile robot localization. However, it is still difficult to guarantee its safety because there are no methods determining reliability for MCL estimate. This paper presents a novel localization framework that enables robust localization, reliability estimation, and quick re-localization, simultaneously. The presented method can be implemented using a similar estimation manner to that of MCL. The method can increase localization robustness to environment changes by estimating known and unknown obstacles while performing localization; however, localization failure of course occurs by unanticipated errors. The method also includes a reliability estimation function that enables a robot to know whether localization has failed. Additionally, the method can seamlessly integrate a global localization method via importance sampling. Consequently, quick re-localization from a failure state can be realized while mitigating noisy influence of global localization. We conduct three types of experiments using wheeled mobile robots equipped with a 2D LiDAR. Results show that reliable MCL that performs robust localization, self-failure detection, and quick failure recovery can be realized.
\end{abstract}

\input{introduction}

\input{related_works}

\input{proposed_method}

\input{implementation}

\input{simulation}

\input{experiments}

\input{conclusion}

\section*{Acknowledgments}

This research was supported by the Japan Science and Technology Agency and KAKENHI 18K13727.

\bibliographystyle{apalike}
\bibliography{jfrExample}

\end{document}

%% file: introduction.tex
\section{Introduction}
\label{sec:introduction}

Reliability is a key factor for realizing safety guarantee of fully autonomous robot systems.
Localization is a fundamental module for autonomous navigation systems.
In this work, we focus on 2D LiDAR-based localization and present a method that enables to make localization more reliable.
To realize reliable localization, we consider that localization must
\begin{itemize}
    \item robustly work in dynamic environments;
    \item immediately detect failure of its estimate;
    \item quickly re-localize the robot pose if failed, i.e., quick failure recovery.
\end{itemize}
The method presented in this work can achieve these functions simultaneously.
It should be noted that we only present 2D-LiDAR-based implementation of the presented framework in this paper, but the framework can be applied to other localization problems such as 3D-LiDAR-based one.

The presented method is composed of two processes; pose tracking and global localization.
The pose tracking process is basically implemented based on Monte Carlo localization (MCL)~\cite{Thrun_PR}.
Additionally, the sensor measurement class and reliability estimation methods we previously presented in~\cite{AkaiIROS2018, AkaiIV2018} are integrated.
The global localization process is implemented based on the free-space feature presented in~\cite{8967683}.
Global localization is necessary to achieve re-localization from failure of pose tracking; however, global localization does not stably work more than pose tracking in usual.
To seamlessly fuse global localization and pose tracking, the probabilistic fusion method we previously presented in~\cite{AkaiICRA2020} is employed.
By using these methods, the above functions can be achieved.

The contribution of this paper is achieving integration of the above mentioned methods.
As mentioned above, the presented method is basically composed of our previous proposals.
However, we did not achieve these integration owing to less of a model and implementation difficulty.
In this work, we present a new graphical model for the pose tracking process and solve the global localization problem using the free-space feature.
As a result, reliable localization that simultaneously achieves the above three things can be realized.
We conduct simulation-, dataset- and our-own-platform-based experiments using wheeled mobile robots equipped with a 2D LiDAR.
Through the simulation-based experiments, we show that the presented method numerically works well more than traditional methods.
Through the dataset- and our-own-platform-based experiments, we show that the presented method also works with actual robots.
The software used in this work is publicly available at\footnote{\url{https://github.com/NaokiAkai/als_ros}}.
The contribution is summarized as follows.
\begin{itemize}
    \item Presenting a new framework that integrates our previous proposals presented in~\cite{AkaiIROS2018, AkaiIV2018, AkaiICRA2020} and achieves reliable localization that simultaneously performs robust localization, reliability estimation, and quick re-localization
    \item Publishing ROS-based implementation of the new framework with 2D LiDAR as open source software
\end{itemize}

The rest of this paper is organized as follows.
Section~\ref{sec:related_work} summarizes related works.
Section~\ref{sec:proposed_method} describes the problem setting and details the presented method.
Section~\ref{sec:implementation} details implementation of the presented method.
Section~\ref{sec:simulation_experiments} and \ref{sec:experiments} describe experimental results using simulation, dataset and our own platform.
Section~\ref{sec:conclusion} concludes this work.

%% file: related_works.tex
\section{Related work}
\label{sec:related_work}

This section summarizes existing works related to robust localization, reliability estimation, and re-localization.

\subsection{Robustness}

To perform localization, we need to model how sensor measurements are obtained.
This model is referred to the measurement model~\cite{Thrun_PR}.
The traditional measurement models are the beam and likelihood field models presented in~\cite{Thrun_PR}.
These models consider that unknown obstacles, that is, obstacles do not exist on the map, are measured.
Hence, these models enable to perform localization in dynamic environments.
However, it is difficult for these models to work in highly dynamic environments because highly dynamic environment yields inconsistency of the models.

The major dynamics in real world are moving obstacles such as pedestrians, bicycles, and cars.
The simplest method to cope with them is to treat them as outliers~\cite{FoxJAIR1999, BurgardAI1999, MontemerloICRA2002, AnguelovUAI2002, HahnelICRA2003, SchulzIJCAI2003}.
In other words, measurements obtained from these obstacles are ignored in the likelihood calculation process.
This approach effectively works for such dynamic obstacles; however, it cannot cope with semi-dynamic obstacles such as parked cars and removal of mapped obstacles.

An other effective approach to cope with dynamic obstacles is to model static and dynamic parts of the environment.
Wolf and Sukhatme~\cite{WolfAR2005} proposed an occupancy grid mapping method that separately maps static and dynamic parts.
Montesano {\it et al}.~\cite{MontesanoICRA2005} also proposed a modeling method of static and dynamic parts.
This modeling is achieved using a set of filters tracking the moving objects and a map of the static structure constructed online.
Dynamic environment modeling methods have been extended by many authors~\cite{BrechtelICRA2010, Meyer-DeliusAAAI2012, SaarinenIROS2012, TipaldiIJRR2013}.
Wang {\it et al}.~\cite{WangIJRR2007} extended the probabilistic method and proposed a general framework for simultaneous localization, mapping, and dynamic object tracking.
Biber and Duckett~\cite{BiberRSS2005} proposed a modeling method that represent multiple timescales to select the sensor measurements of the most appropriate timescale.
Meyer-Delius {\it et al}.~\cite{Meyer-DeliusIROS2010} and Valencia {\it et al}.~\cite{ValenciaICRA2014} also proposed a localization method that uses temporal maps.
These approaches are able to provide a better static map for localization.
However, these approaches require increasing of memory cost for modeling.
The presented method increases localization robustness without increasing memory and computational costs.


Improving the measurement model is also effective to improve localization robustness.
Olufs and Vincze~\cite{OlufsIRSO2009} and Takeuchi {\it et al}.~\cite{TakeuchiROBIO2010} proposed the measurement model that utilizes the free space.
Appropriate sensor measurement selection such as presented by Kim and Chung~\cite{KimICRA2018} is also effective to improve robustness.
However, these extensions are not to extend the graphical model for localization presented in~\cite{Thrun_PR}.
Our approach extends the graphical model and formulates the simultaneous localization and sensor measurement class estimation problem (see Fig.~\ref{fig:graphical_model}).
The basic model used in our approach is different from that of used in these approaches.
Consequently, localization robustness to environment changes can be improved.

Our approach to improve localization robustness is similar to the method presented by Yang and Wang~\cite{YangICRA2011}.
In their method, the feasibility grids that maintain the stochastic estimates of the feasibility (crossability) states of the environment are used.
Based on the feasibility grids, sensor measurements can be decomposed into stationary and moving objects.
Our approaches also uses two class conditional measurement models for known and unknown obstacles; however, no other information is used to calculate them such as the feasibility grids.

\subsection{Reliability}

MCL is a well-known algorithm for mobile robot localization~\cite{DellaertICRA1999} and is based on Bayesian filtering.
MCL does not include a failure detection function in its standard implementation since the Bayesian filter just estimates the posterior over the target variable, i.e., robot pose.
Gutmann and Fox~\cite{gutmann_iros2002:_amcl} proposed a failure detection method for MCL that observes history of likelihood.
MCL including this method is referred to augmented MCL~(AMCL).
Failure detection by AMCL is performed based on thresholds.

Scan registration methods such as iterative closest points (ICP)~\cite{BeslTPAMI1992} and normal distributions transform (NDT)~\cite{BiberIROS2003} are also widely used for mobile robot localization.
These methods also do not have an exact failure detection function in its standard implementation.
The simplest method to detect failures for these algorithms is setting thresholds for distances between corresponding points and/or features~\cite{Rusinkiewicz3DDIMM}.
However, insufficient performance of the threshold-based method when registration errors are small is presented in~\cite{SilvaTPAMI2005}.
In the presented method, the threshold-based method is also used; however, we can improve classification stability because the proposed graphical model derives the Bayesian filter for reliability estimation.

Quddus {\it et al}.~\cite{QuddusTRC2006} presented an empirical methods that use map matching to estimate localization integrity.
In this method, fuzzy logic is used to determine a metric between 0 and 100 that represents uncertainty (confidence) of the map matching results.
However, we consider that confidence and reliability for localization results are different because confidence can be calculated from the pose estimation uncertainty, but reliability cannot be calculated.
Our approach employees a localization classifier and estimates reliability using the classifier.

Sundvall and Jensfelt~\cite{SundvallICRA2006} and Mendoza {\it et al}.~\cite{MendozaCMU2018} proposed a localization failure detection method using a redundant positioning system.
The main idea of these methods is to use majority vote of the redundant system to detect failures.
In other words, these methods do not estimate correctness of each estimate in the redundant system.
Our approach explicitly estimates reliability of the localization system.
Al Hage {\it et al}.~\cite{HageT-ITS2022} proposed a method for addressing localization integrity that combines measurement rejection and position error characterization.
In this method, a multi-sensor data fusion with a fault detection and exclusion algorithm is constituted using a bank of information filters.
Our approach does not use multi-sensor data for estimating reliability.

Modeling localization failure is not trivial.
Hence, machine learning approaches have been applied recently.
Almqvist {\it et al}.~\cite{Almqvist2018LearningTD} applied several threshold- and machine-learning-based methods to classify misaligned point clouds.
Alsayed {\it et al}. ~\cite{AlsayedITSC2017, AlsayedICRA2018} presented a machine-learning-based failure detection method for 2D LiDAR SLAM.
Similar approaches for GNSS-based localization is also presented in~\cite{HsuITSC2017}.
Zhen {\it et al}.~\cite{ZhenICRA2017} presented ``localizability'' that represents possibility whether localization works or not in each area.
The localizability cannot exactly describe localization performance because it is determined using only geometric shape of the map, i.e., it does not consider on-line situation.
Nobili {\it et al}.~\cite{NobiliICRA2018} extended the localizability and proposed ``alignability'' that represents alignment risk.
In addition, support vector machine is applied to learn relationship between the alignability and localization failures.
These methods can indeed detect localization failures; however, these methods cannot consider uncertainty of the classifier.
Our approach can consider the uncertainty because reliability estimation is performed based on the Bayesian filtering using the classification results.

It should be noted that the estimated reliability by the presented method cannot perfectly guarantee localization correctness since it is also estimated without the ground truth.
Interpretation of reliability in this work is discussed in Section~\ref{subsubsec:reliability_estimation_accuracy}.


\subsection{Global localization and re-localization}

MCL can be also used for global localization~\cite{DellaertICRA1999}.
However, standard MCL is not suitable for solving the global localization problem.
To efficiently solve the global localization problem using MCL, there are several extensions.
It should be noted that we treat that the global localization and re-localization problems are the same because these problems aim to estimate the current pose without accurate initial and/or previous estimates.

Lenser and Veloso~\cite{LenserICRA2000} proposed the sensor resettings that enables to perform re-localization.
In the sensor resettings, failure of MCL estimate is detected with the scheme used in AMCL.
If failure is detected, the particles are sampled using the measurement model.
Then, all the particles are evaluated using the measurement model and perform re-localization.
Ueda {\it et al}.~\cite{UedaIROS2004} proposed the expansion resettings that also enables to perform re-localization.
In the expansion resettings, failure of MCL estimate is also detected with the AMCL's scheme.
If failure is detected, the particle distribution is expanded around the current estimate.
The expansion resettings is suitable for recovering small estimation errors caused by such as wheel slippage and collision against obstacles.

Thrun {\it et al}.~\cite{ThrunAI2001} proposed mixture MCL (MMCL) that uses multiple proposal distributions and importance sampling to fuse particles generated from multiple distributions.
In~\cite{ThrunAI2001}, the measurement model is used as the proposal distribution.
Hence, this method is similar to the sensor resettings; however, it differs from the sensor resettings because the fusion is performed based on the importance sampling.
However, accurate sampling using the measurement model is not a trivial task in an era when MMCL has been proposed because of the computational and memory costs.
Thrun {\it et al}.~\cite{ThrunNIPS2001} also proposed a risk sensitive particle filter (RSPF) that considers the risk estimated by Markov decision process.
In~\cite{ThrunNIPS2001}, it was shown that RSPF achieved faster recovery from localization failure more than standard MCL.

Recent evolution of neural networks also contributes to the localization filed.
Kendall {\it et al}.~\cite{KendallICCV2015} proposed PoseNet that directly estimates 6-DoF pose from a camera image, i.e., networks take sensor measurements and infer location where the sensor measurements are obtained and such networks are referred to end-to-end (E2E) networks.
Gal and Ghahramani~\cite{Gal2015BayesianCN} showed that output of neural networks can be treated as posterior by approximating using variational inference.
Kendall and Cipolla~\cite{Kendall2016ModellingUI} used this idea and showed that uncertainty of E2E-based localization can be estimated.
This means that poses can be sampled from the probability over the measurement, i.e., this sampling is similar to sampling from the measurement model.
Sun {\it et al}.~\cite{SunICRA2020} and we~\cite{AkaiICRA2020} used this sampling method and proposed a localization method that fuses MCL and E2E-based localization.
Owing to the fusion, smooth pose tracking and fast re-localization can be simultaneously achieved.

However, the use of E2E networks requires users preparation of training dataset because coordinates of a map depend on how to build a map.
In other words, E2E networks are needed to be trained and this training depends on environment and map building process.
We consider that this requirement is not suitable for the use in open source software.
Hence, we use a model-based method presented in~\cite{8967683}.
This re-localization performance might not overcome the E2E-based one; however, we show that accurate and fast re-localization can be achieved using the fusion based on the importance sampling.
It should be noted that the opened software is implemented to easily change the global localization method, that is, readers can easily change it according to application, for example, GNSS and AR marker.

%% file: proposed_method.tex
\section{Proposed method}
\label{sec:proposed_method}

This section first describes the problem setting.
Then, the pose tracking and global localization methods contained in the presented localization framework are detailed.

\subsection{Problem setting}
\label{subsec:problem_setting}

In this work, we focus on the 2D localization problem for mobile robots.
A pose of a target robot is composed of 2D position, $x$ and $y$, and heading angle, $\theta$.
We assume that a robot is equipped with a 2D LiDAR and an inertial navigation system (INS).
Measurements of INS are denoted as ${\bf u}$ and are used to predict the robot pose.
Measurements of LiDAR are denoted as ${\bf z}$ and are used to perform matching with a map denoted as ${\bf m}$.
In this work, we aim to estimate the current robot pose with sequences of ${\bf u}$ and ${\bf z}$, and ${\bf m}$.
Additionally, we try to realize robust localization, reliability estimation, and quick re-localization simultaneously.
Robust localization and reliability estimation are realized by the improved pose tracking algorithm described in the next subsection.
Quick re-localization is realized by the combination of global localization and the probabilistic fusion of it with the pose tracking algorithm.

\subsection{Graphical model for pose tracking}

\subsubsection{Variable definition}

\begin{figure}[!t]
    \begin{center}
        \includegraphics[width = 70 mm]{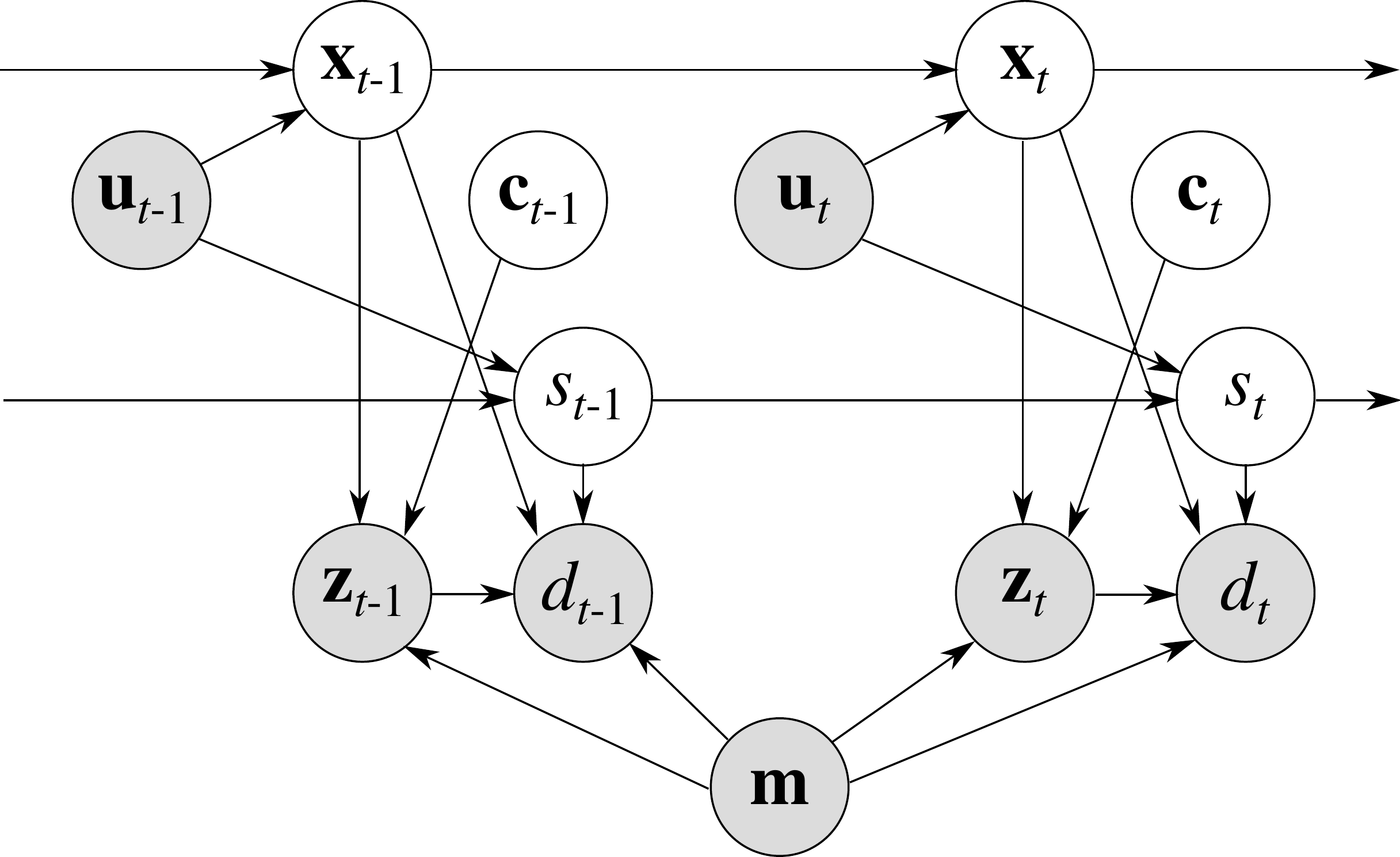}
        \caption{The graphical model for the pose tracking.}
        \label{fig:graphical_model}
    \end{center}
\end{figure}

Figure~\ref{fig:graphical_model} illustrates the graphical model for the pose tracking process.
The white and gray nodes indicate hidden and observable variables.
The robot pose ${\bf x}$, measurement classes ${\bf c}$, and localization state $s$ are treated as the hidden variables.
The INS and LiDAR measurements, ${\bf u}$ and ${\bf z}$, map ${\bf m}$, and output of a localization state classifier $d$ are treated as the observable variables.

The LiDAR measurements are denoted as ${\bf z} = ({\bf z}^{[1]}, {\bf z}^{[2]}, ..., {\bf z}^{[K]})$, where ${\bf z}^{[k]}$ denotes $k$th measurement.
The measurement classes ${\bf c}$ indicate category of the LiDAR measurements and denoted as ${\bf c} = (c^{[1]}, c^{[2]}, ..., c^{[K]})$, where $c^{[k]}$ corresponds to ${\bf z}^{[k]}$.
In this work, we use two classes denoted as $c^{[k]} \in \mathcal{C} = \{ {\rm known}, {\rm unknown} \}$.
These classes indicate whether the measurement is obtained from obstacles existing on the map or not.

The localization state classifier distinguishes whether localization has failed.
The classifier can be implemented using any methods such as threshold- and machine-learning-based methods~\cite{Almqvist2018LearningTD}.
Output form of the classifier is changed according to its implementation.
In this work, we use a threshold-based classifier that sets a threshold to mean absolute error (MAE) defined using residual errors and $d$ is to be an one dimensional continuous value, where the residual errors represent a distance set between LiDAR measurement points and the closest mapped obstacle for each point.
The localization state $s$ is denoted as $s \in \mathcal{S} = \{ {\rm success}, {\rm failure} \}$.
These states indicate whether localization has succeeded.
The state $s$ is estimated using outputs of the classifier $d$.
Since $s = {\rm success}$ indicate that localization has succeeded, reliability can be known by calculating its probability, i.e., $p(s = {\rm success})$.

\subsubsection{Formulation}

Our objective is to estimate the joint posterior distribution over the robot pose, measurement classes, and localization state at current time $t$ as shown in Eq.~(\ref{eq:joint_posterior}).
\begin{align}
    p({\bf x}_{t}, {\bf c}_{t}, s_{t} | {\bf u}_{1:t}, {\bf z}_{1:t}, d_{1:t}, {\bf m}),
    \label{eq:joint_posterior}
\end{align}
where $1:t$ indicates time sequence, i.e., ${\bf u}_{1:t} = ({\bf u}_{1}, {\bf u}_{2}, ..., {\bf u}_{t})$.
Eq.~(\ref{eq:joint_posterior}) can be decomposed using the multiplication theorem
\begin{eqnarray}
    p({\bf x}_{t}, {\bf c}_{t}, s_{t} | {\bf u}_{1:t}, {\bf z}_{1:t}, d_{1:t}, {\bf m}) =
    & p({\bf x}_{t} | {\bf u}_{1:t}, {\bf z}_{1:t}, d_{1:t}, {\bf m}) \label{eq:pose_posterior} ~~~~~~~~~ \\
    & p({\bf c}_{t} | {\bf x}_{t}, {\bf u}_{1:t}, {\bf z}_{1:t}, d_{1:t}, {\bf m}) \label{eq:measurement_classes_posterior} ~~~~~ \\
    & p(s_{t} | {\bf x}_{t}, {\bf c}_{t}, {\bf u}_{1:t}, {\bf z}_{1:t}, d_{1:t}, {\bf m}), \label{eq:localization_state_posterior}
\end{eqnarray}
where Eqs.~(\ref{eq:pose_posterior}), (\ref{eq:measurement_classes_posterior}), and (\ref{eq:localization_state_posterior}) denote the posterior over the robot pose, measurement classes, and localization state, respectively.

Eqs.~(\ref{eq:pose_posterior}), (\ref{eq:measurement_classes_posterior}), and (\ref{eq:localization_state_posterior}) cannot be calculated as it is.
We formulate, respectively, them to enable to calculate.
The Bayes theorem is first applied to Eq.~(\ref{eq:pose_posterior}).
\begin{align}
    \begin{split}
        & p({\bf x}_{t} | {\bf u}_{1:t}, {\bf z}_{1:t}, d_{1:t}, {\bf m}) \\
        = & \frac{p({\bf z}_{t} | {\bf x}_{t}, {\bf u}_{1:t}, {\bf z}_{1:t-1}, d_{1:t-1}, {\bf m}) p(d_{t} | {\bf x}_{t}, {\bf u}_{1:t}, {\bf z}_{1:t}, d_{1:t-1}, {\bf m}) p({\bf x}_{t} | {\bf u}_{1:t}, {\bf z}_{1:t-1}, d_{1:t-1}, {\bf m})}
        {p({\bf z}_{t} | {\bf u}_{1:t}, {\bf z}_{1:t-1}, d_{1:t-1}, {\bf m}) p(d_{t} | {\bf u}_{1:t}, {\bf z}_{1:t}, d_{1:t-1}, {\bf m})} \\
        = & \eta p({\bf z}_{t} | {\bf x}_{t}, {\bf m}) p(d_{t} | {\bf x}_{t}, {\bf z}_{t}, {\bf m}) p({\bf x}_{t} | {\bf u}_{1:t}, {\bf z}_{1:t-1}, d_{1:t-1}, {\bf m})
    \end{split}
    \nonumber
\end{align}
To change from the line 2 to 3, the denominator was re-written as a normalization constant, $\eta$, and D-separation~\cite{Bishop_PRML} was applied to the distributions over ${\bf z}_{t}$ and $d_{t}$ to remove non related conditional variables.
The law of total probability is then applied to each distribution.
\begin{align}
    \begin{split}
        & p({\bf z}_{t} | {\bf x}_{t}, {\bf m}) p(d_{t} | {\bf x}_{t}, {\bf z}_{t}, {\bf m}) p({\bf x}_{t} | {\bf u}_{1:t}, {\bf z}_{1:t-1}, d_{1:t-1}, {\bf m}) \\
        = & \sum_{{\bf c}_{t} \in \mathcal{C}} p({\bf z}_{t} | {\bf x}_{t}, {\bf c}_{t}, {\bf m}) p({\bf c}_{t})
          \sum_{s_{t} \in \mathcal{S}} p(d_{t} | {\bf x}_{t}, s_{t}, {\bf z}_{t}, {\bf m}) p(s_{t}) \\
        & \int p({\bf x}_{t} | {\bf x}_{t-1}, {\bf u}_{1:t}, {\bf z}_{1:t-1}, d_{1:t-1}, {\bf m}) p({\bf x}_{t-1} | {\bf u}_{1:t}, {\bf z}_{1:t-1}, d_{1:t-1}, {\bf m}) {\rm d}{\bf x}_{t-1} \\
        = & \sum_{{\bf c}_{t} \in \mathcal{C}} p({\bf z}_{t} | {\bf x}_{t}, {\bf c}_{t}, {\bf m}) p({\bf c}_{t})
          \sum_{s_{t} \in \mathcal{S}} p(d_{t} | {\bf x}_{t}, s_{t}, {\bf z}_{t}, {\bf m}) p(s_{t}) 
        \int p({\bf x}_{t} | {\bf x}_{t-1}, {\bf u}_{t}) p({\bf x}_{t-1} | {\bf u}_{1:t-1}, {\bf z}_{1:t-1}, d_{1:t-1}, {\bf m}) {\rm d}{\bf x}_{t-1} \\
    \end{split}
    \nonumber
\end{align}
To change from the line 2 to 3, D-separation was applied to the distribution over ${\bf x}_{t}$ and ${\bf u}_{t}$ was omitted from the distribution over ${\bf x}_{t-1}$ since the future input does not affect to the previous pose.
The recursive update equation regarding the pose distribution is obtained.

Eq.~(\ref{eq:measurement_classes_posterior}) can be formulated by applying the Bayes theorem and D-separation.
\begin{align}
    \begin{split}
        p({\bf c}_{t} | {\bf x}_{t}, {\bf u}_{1:t}, {\bf z}_{1:t}, d_{1:t}, {\bf m})
        = & \eta p({\bf z}_{t} | {\bf x}_{t}, {\bf c}_{t}, {\bf u}_{1:t}, {\bf z}_{1:t-1}, d_{1:t}, {\bf m}) p({\bf c}_{t} | {\bf x}_{t}, {\bf u}_{1:t}, {\bf z}_{1:t-1}, d_{1:t}, {\bf m}) \\
        = & \eta p({\bf z}_{t} | {\bf x}_{t}, {\bf c}_{t}, {\bf m}) p({\bf c}_{t})
    \end{split}
    \nonumber
\end{align}
To formulate Eq.~(\ref{eq:localization_state_posterior}), the Bayes theorem and D-separation are first applied.
\begin{align}
    \begin{split}
        p(s_{t} | {\bf x}_{t}, {\bf c}_{t}, {\bf u}_{1:t}, {\bf z}_{1:t}, d_{1:t}, {\bf m})
        = & \eta p(d_{t} | {\bf x}_{t}, {\bf c}_{t}, s_{t}, {\bf u}_{1:t}, {\bf z}_{1:t}, d_{1:t-1}, {\bf m}) p(s_{t} | {\bf x}_{t}, {\bf c}_{t}, {\bf u}_{1:t}, {\bf z}_{1:t}, d_{1:t-1}, {\bf m}) \\
        = & \eta p(d_{t} | {\bf x}_{t}, s_{t}, {\bf z}_{t}, {\bf m}) p(s_{t} | {\bf x}_{t}, {\bf c}_{t}, {\bf u}_{1:t}, {\bf z}_{1:t}, d_{1:t-1}, {\bf m})
    \end{split}
    \nonumber
\end{align}
Then, the law of total probability is applied to the second distribution.
\begin{align}
    \begin{split}
        & p(s_{t} | {\bf x}_{t}, {\bf c}_{t}, {\bf u}_{1:t}, {\bf z}_{1:t}, d_{1:t-1}, {\bf m}) \\
        = & \sum_{s_{t-1} \in \mathcal{S}} p(s_{t} | {\bf x}_{t}, {\bf c}_{t}, s_{t-1}, {\bf u}_{1:t}, {\bf z}_{1:t}, d_{1:t-1}, {\bf m}) p(s_{t-1} | {\bf x}_{t}, {\bf c}_{t}, {\bf u}_{1:t}, {\bf z}_{1:t}, d_{1:t-1}, {\bf m}) \\
        = & \sum_{s_{t-1} \in \mathcal{S}} p(s_{t} | s_{t-1}, {\bf u}_{t}) p(s_{t-1} | {\bf x}_{t-1}, {\bf c}_{t-1}, {\bf u}_{1:t-1}, {\bf z}_{1:t-1}, d_{1:t-1}, {\bf m})
    \end{split}
    \nonumber
\end{align}
To change from the line 2 to 3, D-separation was applied to the distribution over $s_{t}$ and ${\bf x}_{t}$, ${\bf c}_{t}$, ${\bf u}_{t}$, and ${\bf z}_{t}$ was omitted from the distribution over $s_{t-1}$ since the future conditions do not affect to the previous state.
The recursive update equation regarding the localization state is also obtained.

Finally, the target distribution is denoted as shown in Eq.~(\ref{eq:final_joint_posterior}).
\begin{align}
    \begin{split}
        & p({\bf x}_{t}, {\bf c}_{t}, s_{t} | {\bf u}_{1:t}, {\bf z}_{1:t}, d_{1:t}, {\bf m}) \\
        = & \eta \underbrace{ \sum_{{\bf c}_{t} \in \mathcal{C}} p({\bf z}_{t} | {\bf x}_{t}, {\bf c}_{t}, {\bf m}) p({\bf c}_{t})
          \sum_{s_{t} \in \mathcal{S}} p(d_{t} | {\bf x}_{t}, s_{t}, {\bf z}_{t}, {\bf m}) p(s_{t})
          \int p({\bf x}_{t} | {\bf x}_{t-1}, {\bf u}_{t}) p({\bf x}_{t-1} | {\bf u}_{1:t-1}, {\bf z}_{1:t-1}, d_{1:t-1}, {\bf m}) {\rm d}{\bf x}_{t-1} }_{{\rm robot~pose}} \\
          & \underbrace{ p({\bf z}_{t} | {\bf x}_{t}, {\bf c}_{t}, {\bf m}) p({\bf c}_{t}) }_{{\rm measurement~classes}} \\
          & \underbrace{ p(d_{t} | {\bf x}_{t}, s_{t}, {\bf z}_{t}, {\bf m}) \sum_{s_{t-1} \in \mathcal{S}} p(s_{t} | s_{t-1}, {\bf u}_{t}) p(s_{t-1} | {\bf x}_{t-1}, {\bf c}_{t-1}, {\bf u}_{1:t-1}, {\bf z}_{1:t-1}, d_{1:t-1}, {\bf m}) }_{{\rm localization~state}},
    \end{split}
    \label{eq:final_joint_posterior}
\end{align}
%
In Eq.~(\ref{eq:final_joint_posterior}), we have four important models denoted as $p({\bf z}_{t} | {\bf x}_{t}, {\bf c}_{t}, {\bf m})$, $p(d_{t} | {\bf x}_{t}, s_{t}, {\bf z}_{t}, {\bf m})$, $p({\bf x}_{t} | {\bf x}_{t-1}, {\bf u}_{t})$, and $p(s_{t} | s_{t-1}, {\bf u}_{t})$.
These models are referred to the class conditional measurement, decision, motion, and reliability transition models, respectively.
These models are detailed in Section~\ref{subsec:motion_model}, \ref{subsec:reliability_transition_model}, \ref{subsubsec:class_conditional_measurement_model}, and \ref{subsubsec:decision_model}, respectively.

In this work, we use Rao-Blackwellized particle filter (RBPF) to estimate Eq.~(\ref{eq:final_joint_posterior}).
More precisely, Eq.~(\ref{eq:pose_posterior}) is estimated using particle filter (PF) and Eqs.~(\ref{eq:measurement_classes_posterior}) and (\ref{eq:localization_state_posterior}) are estimated using analytical methods based on sampled poses in PF.

\subsection{Global localization and its fusion with pose tracking}

In this work, we assume that the global localization problem is to estimate the probabilistic distribution over the robot pose under a condition where several INS and LiDAR measurements and map are given.
Hence, we also try to estimate the distribution shown in Eq.~(\ref{eq:re-localization}).
\begin{align}
    p({\bf x}_{t} | {\bf u}_{t-N+1:t}, {\bf z}_{t-N:t}, {\bf m}),
    \label{eq:re-localization}
\end{align}
where $N$ is the number of previous steps used for global localization.
It should be noted that the INS measurement ${\bf u}$ is ignored if $N=0$.

As shown in Eqs.~(\ref{eq:pose_posterior}) and (\ref{eq:re-localization}), there are two probabilistic distributions over the robot pose in the presented method.
Ideally, global localization has to be performed only when localization has failed.
However, exact failure detection of localization is challenging owing to the use of the independent assumption to the LiDAR measurements~\cite{Thrun_PR, AkaiRA-L2019, AkaiT-ITS2022}.
Even though the presented method estimates reliability of the localization result, it is difficult to perfectly classify whether localization has failed using the estimated reliability.
Namely, seamless fusion of pose tracking and global localization is preferable more than switching of them according to estimate condition.

To realize the seamless fusion, we use the importance sampling.
First, poses are sampled from Eq.~(\ref{eq:re-localization}).
Let ${}^{\rm G}M$ be a number of sampled poses and ${}^{\rm G}{\bf x}_{t}^{[i]}$ be a $i$th sampled pose, we assume that the sampled poses approximate Eq.~(\ref{eq:re-localization}) like the approximation by PF.
\begin{align}
    p({\bf x}_{t} | {\bf u}_{t-N+1:t}, {\bf z}_{t-N:t}, {\bf m}) \simeq \frac{1}{{}^{\rm G}M} \sum_{i=1}^{{}^{\rm G}M} \delta ({\bf x}_{t} - {}^{\rm G}{\bf x}_{t}^{[i]}),
    \label{eq:re-localization_approximation}
\end{align}
where $\delta(\cdot)$ is the Dirac delta that is 1 if value within the blackett is 0, and 0 otherwise.
By the approximation shown in Eq.~(\ref{eq:re-localization_approximation}), we can use the distribution shown in Eq.~(\ref{eq:re-localization}) as a proposal distribution, i.e., distribution for sampling the particles.

In PF, likelihood of the particles is determined by quotient of the target and proposal distributions.
The target distribution used in the presented method is shown in Eq.~(\ref{eq:pose_posterior}).
Hence, the likelihood of the particles sampled from Eq.~(\ref{eq:re-localization}) is denoted as
\begin{align}
    \begin{split}
        {}^{\rm G}\omega_{t} = & \frac{ p({\bf x}_{t} | {\bf u}_{1:t}, {\bf z}_{1:t}, d_{1:t}, {\bf m}) }{ p({\bf x}_{t} | {\bf u}_{t-N+1:t}, {\bf z}_{t-N:t}, {\bf m}) } \\
        \simeq & \eta \frac{ \sum_{{\bf c}_{t} \in \mathcal{C}} p({\bf z}_{t} | {\bf x}_{t}, {\bf c}_{t}, {\bf m}) p({\bf c}_{t})
          \sum_{s_{t} \in \mathcal{S}} p(d_{t} | {\bf x}_{t}, s_{t}, {\bf z}_{t}, {\bf m}) p(s_{t}) }
        { \frac{1}{{}^{\rm G}M} \sum_{i=1}^{{}^{\rm G}M} \delta ({\bf x}_{t} - {}^{\rm G}{\bf x}_{t}^{[i]}) } \\
        & \cdot \int p({\bf x}_{t} | {\bf x}_{t-1}, {\bf u}_{t}) p({\bf x}_{t-1} | {\bf u}_{1:t-1}, {\bf z}_{1:t-1}, d_{1:t-1}, {\bf m}) {\rm d}{\bf x}_{t-1},
    \end{split}
\end{align}
where $\int p({\bf x}_{t} | {\bf x}_{t-1}, {\bf u}_{t}) p({\bf x}_{t-1} | {\bf u}_{1:t-1}, {\bf z}_{1:t-1}, d_{1:t-1}, {\bf m}) {\rm d}{\bf x}_{t-1}$ is referred to predictive distribution.

In the robot pose distribution shown in Eq.~(\ref{eq:final_joint_posterior}), the predictive distribution is used to sample the particles and their likelihood is calculated using the class conditional measurement and decision models because the predictive distribution is the proposal distribution.
However, in the likelihood calculation for the sampled particles from Eq.~(\ref{eq:re-localization}), the predictive distribution is used.
Owing to that, seamless fusion of pose tracking and global localization can be realized.

There are various methods to achieve sampling from Eq.~(\ref{eq:re-localization}).
In this work, we use localization using the free-space feature presented in~\cite{8967683}.
This localization method is detailed in Section~\ref{subsec:free_space_feature}.

%% file: implementation.tex
\section{Implementation}
\label{sec:implementation}

\begin{figure}[!t]
    \begin{center}
        \includegraphics[width = 85 mm]{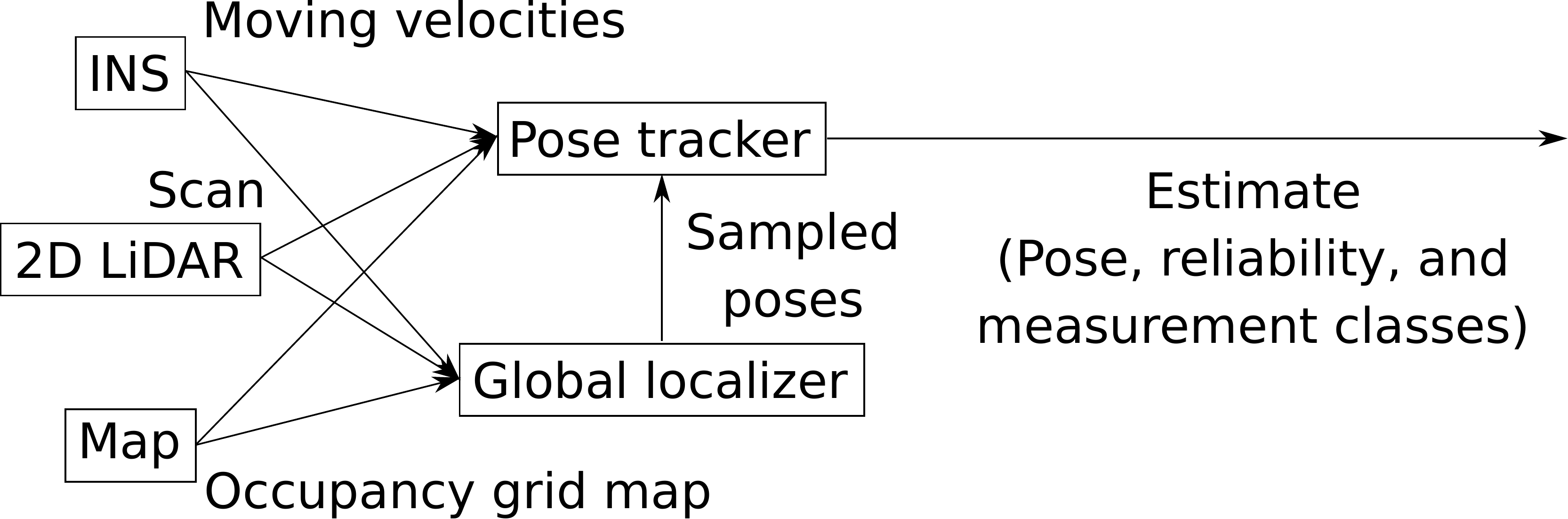}
        \caption{The localization system diagram.}
        \label{fig:system_diagram}
    \end{center}
\end{figure}

The system diagram implemented in this work to realize the functions described in Section~\ref{sec:proposed_method} is illustrated in Fig.~\ref{fig:system_diagram}.
The main modules are the pose tracker and global localizer.
Both the modules receive moving velocities, scan, and occupancy grid map as input.
The global localizer outputs sampled poses based on its estimate.
The pose tracker receives the sampled poses and finally outputs estimated pose.
In addition, the pose tracker outputs reliability to the localization result and measurement classes.

This system is composed of following processes.
\begin{enumerate}
    \item initialization with a given initial pose
    \item update the particles using the motion model
    \item update reliability using the reliability transition model
    \item perform global localization and sample the particles
    \item calculate the likelihood of the particles updated by the motion model using the class conditional measurement and decision models
    \item calculate the likelihood of the particles sampled from global localization using the class conditional measurement and decision models and the predictive distribution
    \item estimate the robot pose, sensor measurement classes, and reliability
    \item re-sampling the particles
    \item go back to 2 and repeat
\end{enumerate}
These processes are detailed in this section.

\subsection{Initialization}

In this work, we focus on the 2D localization problem using a LiDAR.
A state of the particle ${\bf s}_{t}$ contains 2D pose, $x_{t}$, $y_{t}$, and $\theta_{t}$, and likelihood, $\omega_{t}$.
In addition, the particles have reliability denoted as $p(s_{t} = {\rm success})$, where $s_{t}$ is the localization state.
The number of the particles to estimate Eq.~(\ref{eq:pose_posterior}) is constant and is denoted as ${}^{\rm P}M$.
Poses of the particles are randomly sampled around a given initial pose in the initialization step.
Reliability of all the particles is set to $0.5$.
In the presented method, we also have another particle set that is sampled from Eq.~(\ref{eq:re-localization}).
The number of these particles is not constant and is denoted as ${}^{\rm G}M$.

As initialization for the map, we build a distance field~(DF).
Each cell of DF contains distance from the closest mapped obstacle.
DF enables to efficiently calculate the class conditional measurement model.
In addition, DF is used for global localization based on the free-space feature.

\subsection{Motion model}
\label{subsec:motion_model}

We assume that the motion of the robot can be modeled using discrete-time state equation and is denoted as ${\bf x}_{t+1} = f({\bf x}_{t}, {\bf u}_{t})$, where $f(\cdot)$ is the discrete-time state equation.
To update pose of the particles, random noise according to Gaussian is added to the INS measurement ${\bf u}_{t}$.
Update of the particles using the motion model is denoted as
\begin{align}
    {\bf x}_{t}^{[i]} = f({\bf x}_{t-1}^{[i]}, {\bf u}_{t}^{[i]}), ~~~
    {\bf u}_{t}^{[i]} \sim \mathcal{N}({\bf u}_{t}, {}^{\rm u}\Sigma({\bf u}_{t})),
    \label{eq:motion_model}
\end{align}
where ${}^{\rm u}\Sigma$ is a diagonal determined based on the INS measurements, that is, values of the diagonal are to be large (or small) if the INS measurements are large (or small).
In the opened software, the differential drive and omni directional models are supported.

\subsection{Reliability transition model}
\label{subsec:reliability_transition_model}

Basically, localization accuracy decreases according to move of the robot.
Hence, we assume that reliability also decrease according to the move, i.e., $\hat{r}_{t} \leq r_{t-1}$, where $\hat{r}_{t}$ is updated reliability using the reliability transition model from the previous step.
However, modeling the decrease of reliability is not trivial.
We used a heuristical method to model the decrease as shown in Eq.~(\ref{eq:reliability_transition_model}).
\begin{align}
    \hat{r}_{t} = r_{t-1} (1 - \sum_{i} \alpha_{i} \Delta_{i}^{2}),
    \label{eq:reliability_transition_model}
\end{align}
where $\alpha_{i}$ is a positive arbitrary constant and $\Delta_{i}$ is displacement between the time steps measured by INS, for example, translational and angular displacements, $\Delta d$ and $\Delta \theta$, are used if the differential driving model is used.

In the implementation, we set all $\alpha_{i}$ are 0 because effect of the decision model in reliability update is major more than that of the reliability transition model.
In addition, the decision-model-based update can be used in every estimation step.
Hence, effect of the reliability transition model is minor if update cycle is fast, e.g., more than 10~${\rm Hz}$.
These parameters must be appropriately tuned if the update cycle is slow.
Owing to the reliability transition model, we can model that the localization result will be unreliable when sensor measurements are not applied to update its estimate long time.

\subsection{Global localization}

For global localization, we use the free-space feature presented in~\cite{8967683}.
In this subsection, we briefly describe the free-space feature.
For more details, please see the literature.

\subsubsection{Free-space feature}
\label{subsec:free_space_feature}

Let ${\bf m}$ be an occupancy grid map (OGM) and is denoted as ${\bf m} = (m^{[1]}, m^{[2]}, ..., m^{[M]})$, where $m^{i} \in \{ {\rm occupied}, {\rm free~space}, {\rm unknown} \}$ is a state of $i$th cell.
DF is built based on OGM.
A Gaussian filter is applied to DF and a Hessian matrix is calculated in each cell.
Based on the Hessian matrices, maxima, minima, and saddles are detected and these are used as keypoints.
It should be noted that the keypoints are only defined on the free space.

A rotational invariant feature is assigned to the keypoints.
First, a dominant orientation of a keypoint is determined.
A 36-bin gradient orientation histogram is built within an arbitrary size window.
The direction of the maximum frequency value is determined as the dominant orientation of the keypoint.
Then, a 17-bin histogram regarding relative gradient orientations to the dominant orientation is built.
This histogram is assigned to the keypoint as a feature.
In addition, a type of keypoint, i.e., maxima, minima, and saddles, and an average value of DF values in the window are assigned as a feature.

In the matching process of the keypoints, the types of keypoints and average DF values are first compared.
If the types are the same and difference between the average DF values is less than a threshold, these keypoints are regarded as corresponding.
If multiple correspondences are found, the sums of differences of the relative gradient orientations histograms are computed.
If constant multiplication of the minimum sum is less than the second minimum sum, the keypoint with the minimum sum is regarded as corresponding.

\subsubsection{Pose sampling}
\label{subsubsec:pose_sampling}

The free-space features described in Section~\ref{subsec:free_space_feature} are defined to the global map ${\bf m}$ in advance.
In the localization phase, the LiDAR measurements are accumulated based on the INS measurements and a local map ${}^{\rm L}{\bf m}$ is built.
In other words, the local map is built on the odometry frame.
The pose of the odometry frame is denoted as ${}^{{\rm O}}{\bf x}_{t} = ({}^{{\rm O}}x_{t}, {}^{{\rm O}}y_{t}, {}^{{\rm O}}\theta_{t})^{\top}$
The free space features are also defined on the local map and these features are compared with that of the global map.
If corresponding features are found, candidate poses for global localization are determined.

\begin{figure}[!t]
    \begin{center}
        \includegraphics[width = 85 mm]{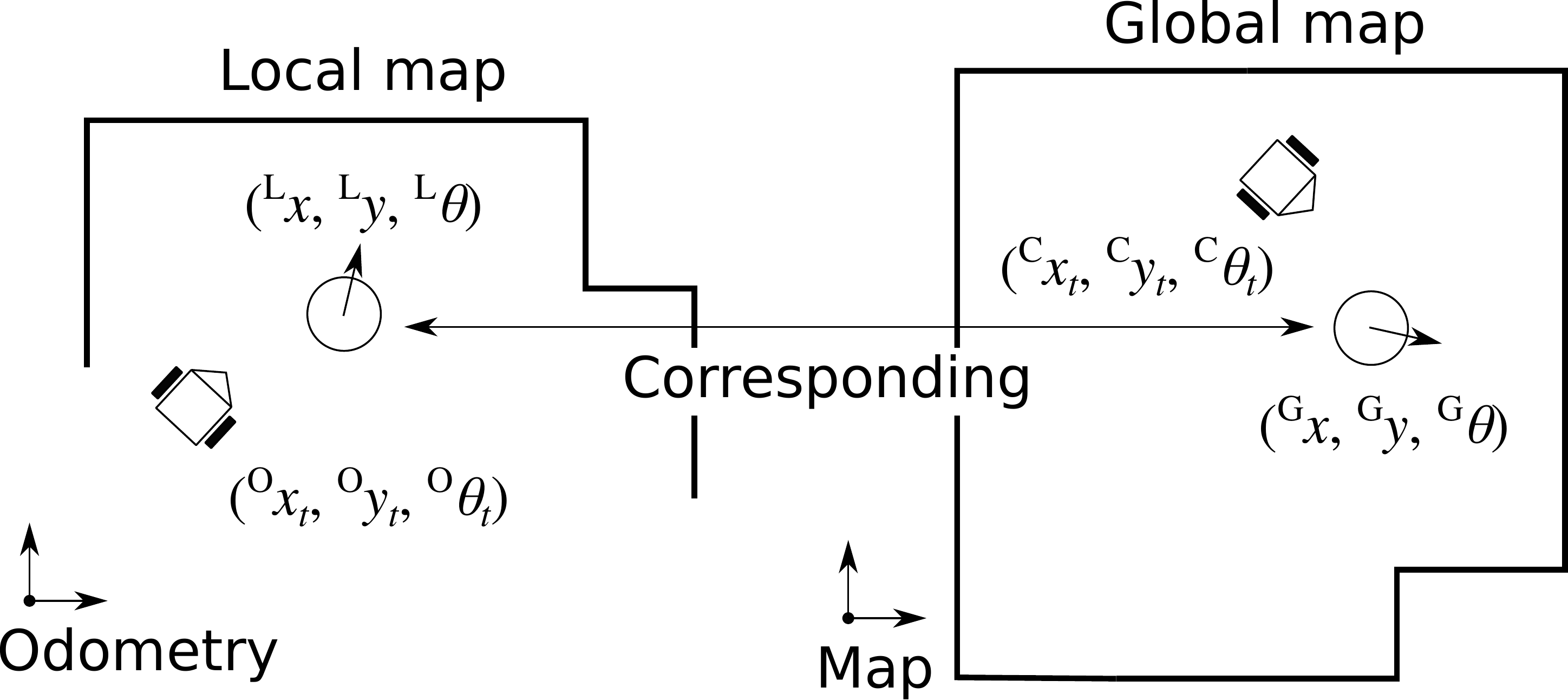}
        \caption{Pose sampling scheme from the global localizer.}
        \label{fig:gl_pose_sampling}
    \end{center}
\end{figure}

Figure~\ref{fig:gl_pose_sampling} illustrates a pose sampling scheme.
Since the free-space feature has dominant orientation, these orientations are matched first.
Then, a position on the map coordinates is determined.
Candidate position and heading direction ${}^{{\rm C}}{\bf x}_{t} = ({}^{{\rm C}}x_{t}, {}^{{\rm C}}y_{t}, {}^{{\rm C}}\theta_{t})^{\top}$ can be determined as follows
\begin{align}
    \begin{gathered}
        \Delta \theta_{1} = {}^{{\rm L}}\theta - {}^{{\rm G}}\theta,~~~
        \Delta \theta_{2} = {}^{{\rm L}}\theta - {}^{{\rm O}}\theta_{t} \\
        \Delta x = {}^{{\rm L}}x - {}^{{\rm O}}x_{t},~~~
        \Delta y = {}^{{\rm L}}y - {}^{{\rm O}}y_{t} \\
        {}^{{\rm C}}x_{t} = \Delta x \cos(\Delta \theta_{1}) - y \sin(\Delta \theta_{1}) + {}^{{\rm G}}x \\
        {}^{{\rm C}}y_{t} = \Delta x \sin(\Delta \theta_{1}) + y \cos(\Delta \theta_{1}) + {}^{{\rm G}}y \\
        {}^{{\rm C}}\theta_{t} = {}^{{\rm G}}\theta - \Delta \theta_{2},
    \end{gathered}
    \label{eq:gl_candidate_pose}
\end{align}
where ${\rm G}$ and ${\rm L}$ indicate the free-space features defined on the global and local maps, and $x$, $y$, and $\theta$ of them indicate the feature position and dominant orientation, respectively.

However, it is difficult to exactly determine the robot pose using Eq.~(\ref{eq:gl_candidate_pose}) since the dominant orientation is roughly determined.
Hence, poses for global localization are sampled by adding random noises to the candidate pose ${}^{{\rm C}}{\bf x}_{t}$.
In addition, matching rate of the LiDAR measurements and map is computed using the sampled poses.
If the matching rate is less than a threshold, the sampled pose is rejected.
Furthermore, we found that the sampled poses sometimes face to opposite side to the ground truth pose.
Hence, we also add opposite directional samples to the candidate poses.

\subsection{Likelihood calculation}
\label{subsec:likelihood_calculation}

\subsubsection{Class conditional measurement model}
\label{subsubsec:class_conditional_measurement_model}

The likelihood of the particles used for estimating $p({\bf x}_{t} | {\bf u}_{1:t}, {\bf z}_{1:t}, d_{1:t}, {\bf m})$, is calculated using two likelihood distributions denoted as $\sum_{{\bf c}_{t} \in \mathcal{C}} p({\bf z}_{t} | {\bf x}_{t}, {\bf c}_{t}, {\bf m}) p({\bf c}_{t})$ and $\sum_{s_{t} \in \mathcal{S}} p(d_{t} | {\bf x}_{t}, s_{t}, {\bf z}_{t}, {\bf m}) p(s_{t})$.
This sub-subsection describes how to calculate the first one.
The second one is described in the next sub-subsection.

We first apply the independent assumption to the LiDAR measurements and decompose the likelihood distribution as shown in Eq.~(\ref{eq:decomposed_ccmm}).
\begin{align}
    \sum_{{\bf c}_{t} \in \mathcal{C}} p({\bf z}_{t} | {\bf x}_{t}, {\bf c}_{t}, {\bf m}) p({\bf c}_{t}) = 
    \prod_{k=1}^{K} \sum_{{\bf c}_{t} \in \mathcal{C}} p({\bf z}_{t}^{[k]} | {\bf x}_{t}, c_{t}^{[k]}, {\bf m}) p(c_{t}^{[k]}),
    \label{eq:decomposed_ccmm}
\end{align}
where $K$ is the number of the LiDAR measurements.
In this work, we consider two sensor measurement classes, i.e., $\mathcal{C} = \{ {\rm known}, {\rm unknown} \}$, where ${\rm known}$ and ${\rm unknown}$ mean whether obstacles exist or do not exist on a given map.
Namely, occupied cells are known obstacles and measurements obtained from other areas are unknown obstacles.
We need to model two class conditional measurement models.
It should be noted that the prior is uniformly set, i.e., $p(c_{t}^{[k]} = {\rm known}) = p(c_{t}^{[k]} = {\rm unknown}) = 0.5$, because there are no available information to estimate the prior.

When the ${\rm known}$ condition is given, the class conditional measurement model is implemented using the likelihood field model~\cite{Thrun_PR}.
%
\begin{align}
    p({\bf z}_{t}^{[k]} | {\bf x}_{t}, c_{t}^{[k]} = {\rm known}, {\bf y}_{t}, {\bf m}) = p_{\rm LFM}({\bf z}_{t}^{[k]} | {\bf x}_{t}, {\bf m})
    = \left(
    \begin{array}{c}
        z_{\rm hit} \\
        z_{\rm max} \\
        z_{\rm rand}
    \end{array}
    \right)^{\top}
    \cdot \left(
    \begin{array}{c}
        p_{\rm hit}({\bf z}_{t}^{[k]} | {\bf x}_{t}, {\bf m}) \\
        p_{\rm max}({\bf z}_{t}^{[k]} | {\bf x}_{t}, {\bf m}) \\
        p_{\rm rand}({\bf z}_{t}^{[k]} | {\bf x}_{t}, {\bf m})
    \end{array}
    \right),
    \label{eq:lfm}
\end{align}
where $z_{\rm hit}$, $z_{\rm max}$, and $z_{\rm rand}$ are arbitrary constants satisfying $z_{\rm hit} + z_{\rm max} + z_{\rm rand} = 1$ and $p_{{\rm hit}}(\cdot)$, $p_{{\rm max}}(\cdot)$, and $p_{{\rm rand}}(\cdot)$ are the measurement models related to the measurement of the known obstacles, maximum value, and random noise, respectively.
The likelihood field model does not explicitly consider measuring dynamic obstacles ($p_{{\rm max}}(\cdot)$ and $p_{{\rm rand}}(\cdot)$ consider measuring noises).
However, the use of the likelihood field model when the ${\rm known}$ condition is given is adequate because the given condition strongly restricts that the measurement is obtained from mapped obstacles.

When the ${\rm unknown}$ condition is given, the class conditional measurement model is implemented using the exponential distribution.
\begin{align}
    p({\bf z}_{t}^{[k]} | {\bf x}_{t}, c_{t}^{[k]}={\rm unknown}, {\bf m}) = \frac{\lambda \exp(-\lambda r_{t}^{[k]})}{1 - \exp(-\lambda r_{\rm max})}.
    \label{eq:exponential_distribution}
\end{align}
where $\lambda$ is the hyperparameter and $r_{\rm max}$ and $r_{t}^{[k]}$ are the maximum measurement and $k$th LiDAR measurement range.
Predicting existence of unknown obstacles is difficult if we do not have any information.
Hence, in the implementation, we assume that unknown obstacles are equally measured within the measurable range.

Eq.~(\ref{eq:decomposed_ccmm}) is calculated as a sum of Eqs.~(\ref{eq:lfm}) and (\ref{eq:exponential_distribution}).
Eqs.~(\ref{eq:lfm}) and (\ref{eq:exponential_distribution}) can be quickly calculated and Eq.~(\ref{eq:exponential_distribution}) can be calculated without further information from that of used in the likelihood field model.
Hence, calculation of the class conditional measurement model does not increase of computational and memory costs.

\subsubsection{Decision model}
\label{subsubsec:decision_model}

Before describing the decision model, we first describe the localization state classifier.
The classifier distinguishes whether localization has failed.
To implement the classifier, mean absolute error (MAE) of residual errors is used in this study.
It should be noted that the implementation is not limited with MAE.
The residual errors are denoted as ${\bf e}_{t} = (e_{t}^{[1]}, e_{t}^{[2]}, ..., e_{t}^{[K]})^{\top}$, where $e_{t}^{[k]}$ is distance from $k$th LiDAR measurement point to the closest mapped obstacle.
MAE is calculated using Eq.~(\ref{eq:mean_abusolute_error}).
\begin{align}
    {\rm MAE}_{t} = \frac{1}{\sum_{k=1}^{K} \mathds{1}(e_{t}^{[k]} \leq e_{\rm max})} \sum_{k=1}^{K} \mathds{1}(e_{t}^{[k]} \leq e_{\rm max}) e_{t}^{[k]},
    \label{eq:mean_abusolute_error}
\end{align}
where $e_{\rm max}$ is the maximum residual error and $\mathds{1}(\cdot)$ is an indicator function which is equal to 1 when the condition within the bracket is true, and 0 otherwise.
We set a threshold to MAE and classify as that localization has failed if MAE exceeds the threshold.
In this implementation, ${\rm MAE}_{t}$ is equal to $d_{t}$ in the graphical model shown in Fig.~\ref{fig:graphical_model}.

The decision model denoted as $p(d_{t} | {\bf x}_{t}, s_{t}, {\bf z}_{t}, {\bf m})$ is needed to be modeled for two cases where localization has succeeded and failed.
Here, we consider that success and failure localization results are positive and negative cases.
In addition, the decision model is composed of true and false classification cases.
Hence, following four distributions are used to calculate the decision model.
\begin{eqnarray}
    p_{\rm true}(d_{t} \leq d_{\rm th} | {\bf x}_{t}, s_{t} = {\rm success}, {\bf z}_{t}, {\bf m}), \label{eq:dm_true_positive} \\
    p_{\rm false}(d_{t} \leq d_{\rm th} | {\bf x}_{t}, s_{t} = {\rm failure}, {\bf z}_{t}, {\bf m}), \label{eq:dm_false_positive} \\
    p_{\rm true}(d_{t} > d_{\rm th} | {\bf x}_{t}, s_{t} = {\rm failure}, {\bf z}_{t}, {\bf m}), \label{eq:dm_true_negative} \\
    p_{\rm false}(d_{t} > d_{\rm th} | {\bf x}_{t}, s_{t} = {\rm success}, {\bf z}_{t}, {\bf m}), \label{eq:dm_false_negative}
\end{eqnarray}
where Eqs.~(\ref{eq:dm_true_positive}), (\ref{eq:dm_false_positive}), (\ref{eq:dm_true_negative}), and (\ref{eq:dm_false_negative}) are the probabilistic distributions over the true positive, false positive, true negative, and false negative cases, respectively.
Note that $\int \left( p_{\rm true}(d_{t} \leq d_{\rm th} | {\bf x}_{t}, s_{t} = {\rm success}, {\bf z}_{t}, {\bf m}) + p_{\rm false}(d_{t} > d_{\rm th} | {\bf x}_{t}, s_{t} = {\rm success}, {\bf z}_{t}, {\bf m}) \right) {\rm d}d_{t} = 1$ and $\int \left( p_{\rm true}(d_{t} > d_{\rm th} | {\bf x}_{t}, s_{t} = {\rm failure}, {\bf z}_{t}, {\bf m}) + p_{\rm false}(d_{t} \leq d_{\rm th} | {\bf x}_{t}, s_{t} = {\rm failure}, {\bf z}_{t}, {\bf m}) \right) {\rm d}d_{t} = 1$.

\begin{figure}[!t]
    \centering
    \subfloat[$p(d_{t} | {\bf x}_{t}, s_{t} = {\rm success}, {\bf z}_{t}, {\bf m})$]{\includegraphics[clip, width = 80 mm]{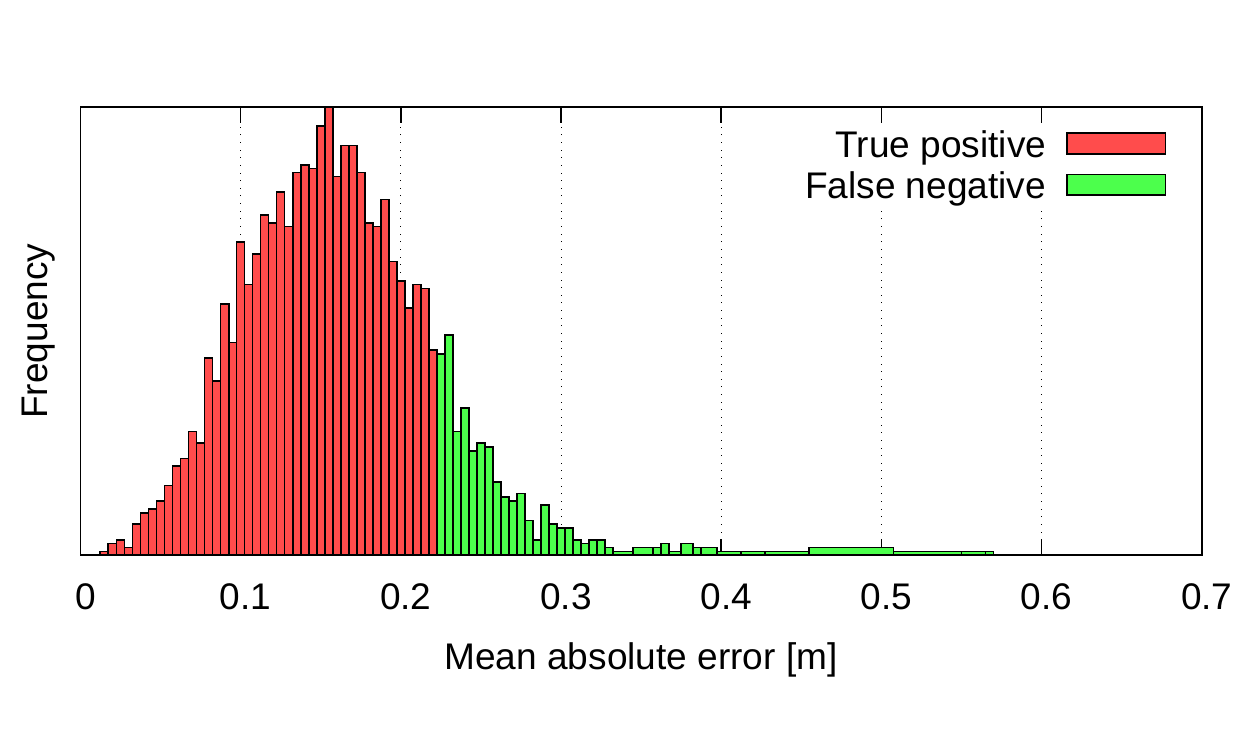}
    \label{fig:decision_model_success}}
    \subfloat[$p(d_{t} | {\bf x}_{t}, s_{t} = {\rm failure},  {\bf z}_{t}, {\bf m})$]{\includegraphics[clip, width = 80 mm]{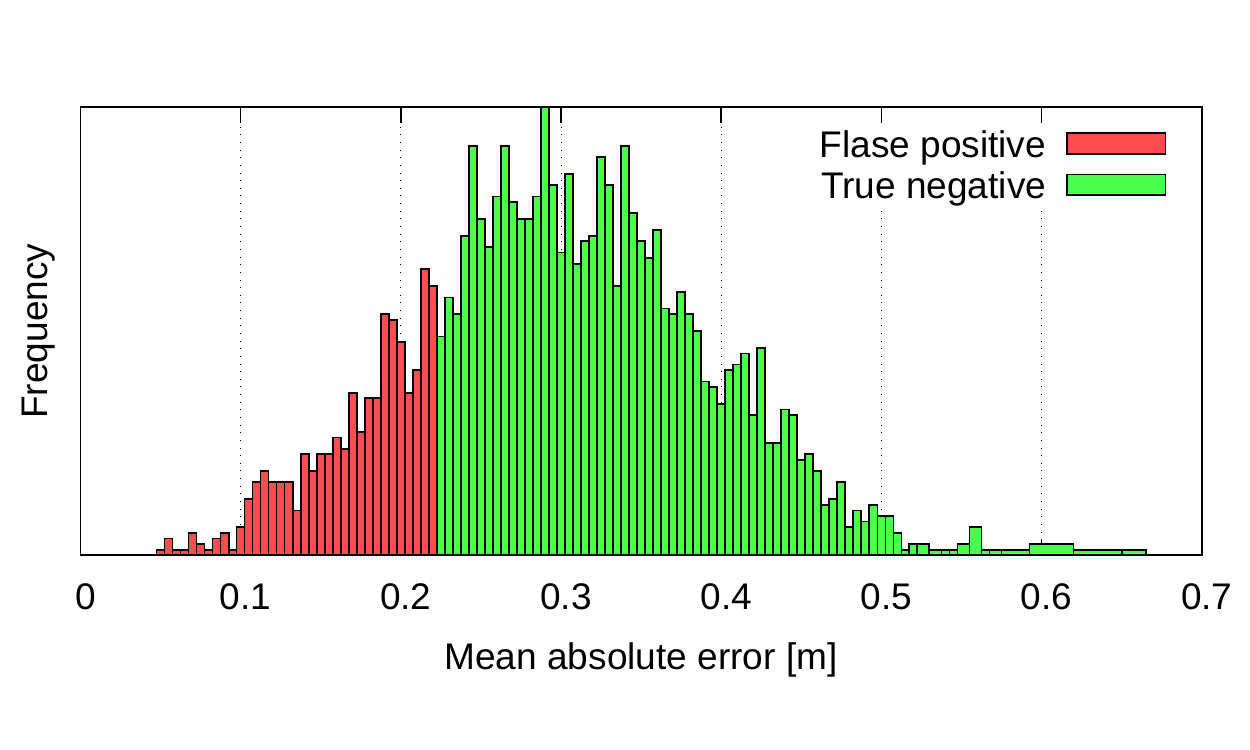}
    \label{fig:decision_model_failure}} \\
    \caption{The decision models of the success (a) and failure (b) cases.}
    \label{fig:decision_models}
\end{figure}

Figure~\ref{fig:decision_models} shows the decision models modeled using histogram.
In this case, we built the dataset using the 2D LiDAR simulation\footnote{\url{https://github.com/NaokiAkai/AutoNavi}}.
Because we used the simulation, the ground truth pose can be obtained.
We aided noises to the ground truth pose and made success and failure poses.
We set positional and angular thresholds to $0.02~{\rm m}$ and $2~{\rm degrees}$ and poses which exceeded either thresholds were classified as failure poses.
The ground truth pose is not used as success poses because it does not have any error.
The threshold to MAE, ${\rm MAE}_{\rm TH}$, was set to $0.225~{\rm m}$ that performed most accurate classification on the dataset.

\begin{figure}[!t]
    \begin{center}
        \includegraphics[width = 85 mm]{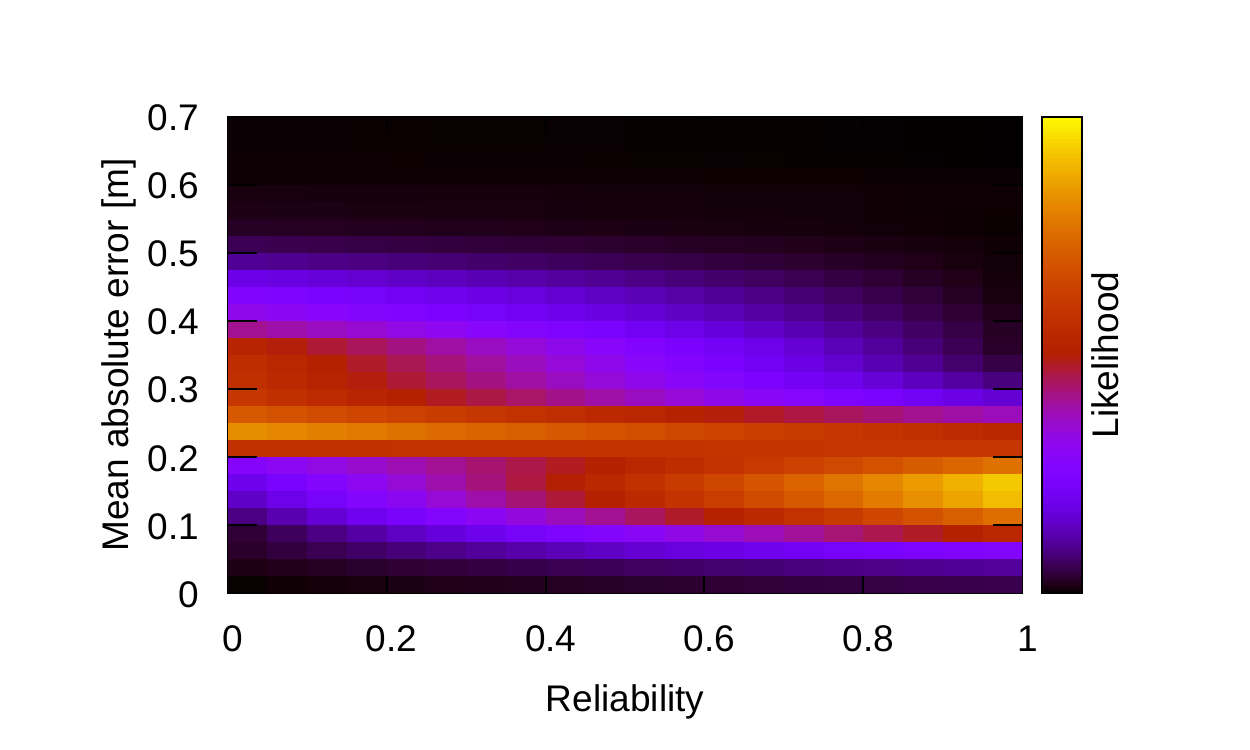}
        \caption{Likelihoods calculated using the decision model.}
        \label{fig:decision_likelihoods}
    \end{center}
\end{figure}

Likelihood calculation results by $\sum_{s_{t} \in \mathcal{S}} p(d_{t} | {\bf x}_{t}, s_{t}, {\bf z}_{t}, {\bf m}) p(s_{t})$ are shown in Fig.~\ref{fig:decision_likelihoods}.
As can be seen from the figure, likelihood is not to be 0 when ${\rm MAE}$ is large (or small) even if reliability is close to 1 (or 0).
Consequently, uncertainty of MAE-based classification can be dealt with.

\subsubsection{Predictive-distribution-based calculation}

The likelihood of the particles sampled from global localization using the free-space feature is calculated using the predictive distribution denoted as $\int p({\bf x}_{t} | {\bf x}_{t-1}, {\bf u}_{t}) p({\bf x}_{t-1} | {\bf u}_{1:t-1}, {\bf z}_{1:t-1}, d_{1:t-1}, {\bf m}) {\rm d}{\bf x}_{t-1}$.
In this sub-subsection, we denote the predictive distribution as $p_{\rm pred}({\bf x}_{t})$.
In PF, the predictive distribution is approximated by the particles updated using the motion model.
\begin{align}
    p_{\rm pred}({\bf x}_{t}) \simeq \sum_{i=1}^{{}^{\rm P}M} \delta({\bf x}_{t} - {\bf x}_{t}^{[i]}),
    \label{eq:approximated_predictive_distribution}
\end{align}
where ${}^{\rm P}M$ is the number of the particles used for pose tracking.
To realize stable localization, likelihood distribution should be smooth; however, the distribution shown in Eq.~(\ref{eq:approximated_predictive_distribution}) is discrete.
Hence, we assume that Eq.~(\ref{eq:approximated_predictive_distribution}) can be approximated by a Gaussian mixture model as shown in Eq.~(\ref{eq:predictive_distribution_gmm}).
\begin{align}
    \sum_{i=1}^{{}^{\rm P}M} \delta({\bf x}_{t} - {\bf x}_{t}^{[i]})
    \simeq \frac{1}{{}^{\rm P}M} \sum_{i=1}^{{}^{\rm P}M} \mathcal{N} \left( {\bf x}_{t}; {\bf x}_{t}^{[i]}, {}^{\rm P} \Sigma \right),
    \label{eq:predictive_distribution_gmm}
\end{align}
where ${}^{\rm P}\Sigma$ is an arbitrary covariance matrix.

However, likelihood of particles that sampled at far from the particles updated using the motion model, ${\bf x}_{t}^{[i]}$, is to be zero if the predictive distribution is modeled using Eq.~(\ref{eq:predictive_distribution_gmm}).
In other words, large localization error cannot be compensated.
To overcome this problem, we model the predictive distribution as shown in Eq.~(\ref{eq:predictive_distribution_global}).
\begin{align}
    p_{\rm pred}({\bf x}_{t})
    \simeq \beta \frac{1}{{}^{\rm P}M} \sum_{i=1}^{{}^{\rm P}M} \mathcal{N} \left( {\bf x}_{t}; {\bf x}_{t}^{[i]}, {}^{\rm P}\Sigma \right)
    + (1-\beta) {\rm unif}({\bf m}),
    \label{eq:predictive_distribution_global}
\end{align}
where $\beta$ is a positive constant including from $0$ to $1$ and ${\rm unif}({\bf m})$ is uniform distribution defined on areas where the robot can exist, that is, the free space.
In the implementation, ${\rm unif}({\bf m})$ is approximated with a small constant value.

\subsubsection{Likelihoods}

In the presented localization method, there are two particle sets.
The first set is used for pose tracking and the second one is sampled from global localization using the free-space feature.
We denote each the particle set as ${}^{\rm P}{\bf s}_{t}^{[i]} = ({}^{\rm P}{\bf x}_{t}^{[i]}, {}^{\rm P}\omega_{t}^{[i]})$ $(i = 1, 2, ..., {}^{\rm P}M)$ and ${}^{\rm G}{\bf s}_{t}^{[i]} = ({}^{\rm G}{\bf x}_{t}^{[i]}, {}^{\rm G}\omega_{t}^{[i]})$ $(i = 1, 2, ..., {}^{\rm G}M)$.
The likelihoods of these particles are calculated using Eqs.~(\ref{eq:likelihood_position_tracking}) and (\ref{eq:likelihood_global_localization}).
\begin{align}
    {}^{\rm P}\omega_{t}^{[i]} = \eta \sum_{{\bf c}_{t} \in \mathcal{C}} p({\bf z}_{t} | {}^{\rm P}{\bf x}_{t}^{[i]}, {\bf c}_{t}, {\bf m}) p({\bf c}_{t}) \sum_{{}^{\rm P}s_{t}^{[i]} \in \mathcal{S}} p(d_{t} | {}^{\rm P}{\bf x}_{t}^{[i]}, {}^{\rm P}s_{t}^{[i]}, {\bf z}_{t}, {\bf m}) p({}^{\rm P}s_{t}^{[i]}),
    \label{eq:likelihood_position_tracking}
\end{align}
\begin{align}
    {}^{\rm G}\omega_{t}^{[i]} = \eta
    \frac{ \sum_{{\bf c}_{t} \in \mathcal{C}} p({\bf z}_{t} | {}^{\rm G}{\bf x}_{t}^{[i]}, {\bf c}_{t}, {\bf m}) p({\bf c}_{t}) \sum_{{}^{\rm G}s_{t}^{[i]} \in \mathcal{S}} p(d_{t} | {}^{\rm G}{\bf x}_{t}^{[i]}, {}^{\rm G}s_{t}^{[i]}, {\bf z}_{t}, {\bf m}) p({}^{\rm G}s_{t}^{[i]}) }
    { \frac{1}{{}^{\rm G}M} \sum_{j=1}^{{}^{\rm G}M} \delta({}^{\rm G}{\bf x}_{t}^{[i]} - {}^{\rm G}{\bf x}_{t}^{[j]}) }
    \cdot p_{\rm pred}({}^{\rm G}{\bf x}_{t}^{[i]}).
    \label{eq:likelihood_global_localization}
\end{align}
It should be noted that the particles sampled from global localization do not have past information.
Hence, the probability regarding the localization state is set to uniform, i.e., $p({}^{\rm G}s_{t}^{[i]} = {\rm success}) = p({}^{\rm G}s_{t}^{[i]} = {\rm failure}) = 0.5$.

\subsection{Estimation}

\subsubsection{Robot pose}

The robot pose is estimated as weight average of the two particle sets.
\begin{align}
    {\bf x}_{t} = 
    \sum_{i=1}^{{}^{\rm P}M} {}^{\rm P}\omega_{t}^{[i]} {}^{\rm P}{\bf x_{t}}^{[i]}
    + \sum_{i=1}^{{}^{\rm G}M} {}^{\rm G}\omega_{t}^{[i]} {}^{\rm G}{\bf x_{t}}^{[i]}.
    \label{eq:pose_estimation}
\end{align}
The likelihoods are normalized before computing Eq.~(\ref{eq:pose_estimation}), i.e., $\sum_{i=1}^{{}^{\rm P}M} {}^{\rm P}\omega_{t}^{[i]} + \sum_{i=1}^{{}^{\rm G}M} {}^{\rm G}\omega_{t}^{[i]} = 1$.

\subsubsection{Sensor measurement classes}

Through the likelihood calculation process, the particle with the maximum likelihood is extracted.
We denote its pose as ${}^{\rm ML}{\bf x}_{t}$.
The probabilistic distribution over the sensor measurement classes is calculated as shown in Eq.~(\ref{eq:estimated_sensor_measurement_classes}).
\begin{align}
    \eta p({\bf z}_{t} | {}^{\rm ML}{\bf x}_{t}, {}^{\rm ML}{\bf c}_{t}, {\bf m}) p({}^{\rm ML}{\bf c}_{t}).
    \label{eq:estimated_sensor_measurement_classes}
\end{align}
This can be calculated using Eqs.~(\ref{eq:lfm}) and (\ref{eq:exponential_distribution}) and these can be quickly calculated.
Hence, this calculation is performed after the likelihood calculation to reduce the memory cost.
A measurement satisfying $p({}^{\rm ML}c_{t}^{[k]} = {\rm unknown}) > \chi$ is detected as an unknown obstacle, where $\chi$ is an arbitrary threshold including from 0 to 1.

\subsubsection{Reliability}

To estimate the reliability, we also use the maximum likelihood particle.
The probabilistic distribution over the reliability is calculated as shown in Eq.~(\ref{eq:estimated_reliability}).
\begin{align}
    \eta p(d_{t} | {}^{\rm ML}{\bf x}_{t}, {}^{\rm ML}s_{t}, {\bf z}_{t}, {\bf m}) p({}^{\rm ML}s_{t}).
    \label{eq:estimated_reliability}
\end{align}
$\eta p(d_{t} | {}^{\rm ML}{\bf x}_{t}, {}^{\rm ML}s_{t} = {\rm success}, {\bf z}_{t}, {\bf m}) p({}^{\rm ML}s_{t} = {\rm success})$ is regarded as the estimated reliability.

\subsection{Re-sampling}

To perform re-sampling, we first calculate values shown in Eq.~(\ref{eq:re-sampling_values}).
\begin{equation}
    b^{[i]} =
    \begin{cases}
        \sum_{j=1}^{i} {}^{\rm P}\omega_{t}^{[j]}                                                                   & \text{if $i \leq {}^{\rm G}M$}, \\
        \sum_{j=1}^{{}^{\rm G}M} {}^{\rm P}\omega_{t}^{[j]} + \sum_{j={}^{\rm G}M+1}^{i} {}^{\rm P}\omega_{t}^{[j - {}^{\rm G}M]} & \text{otherwise}.
    \end{cases}
    \label{eq:re-sampling_values}
\end{equation}
Note that $0 \leq b^{[i]} \leq 1$ and $(i=1, 2, ..., {}^{\rm G}M + {}^{\rm P}M)$.
Then, random values including from 0 to 1, $r^{[i]}$, are generated.
Re-sampling is performed based on Eq.~(\ref{eq:re-sampling}).
\begin{equation}
    {\bf x}_{t}^{[i]} \leftarrow
    \begin{cases}
        {}^{\rm P}{\bf x}_{t}^{[j]}               & \text{if $r^{[i]} \leq b^{[j]}$ and $j \leq {}^{\rm G}M$}, \\
        {}^{\rm G}{\bf x}_{t}^{[j - {}^{\rm G}M]} & \text{if $r^{[i]} \leq b^{[j]}$ and $j > {}^{\rm G}M$}.
    \end{cases}
    \label{eq:re-sampling}
\end{equation}
In addition, likelihood is uniformly reset after re-sampling, i.e., ${}^{\rm P}\omega_{t}^{[i]} = {}^{\rm G}\omega_{t}^{[i]} = 1 / ({}^{\rm G}M + {}^{\rm P}M)$.
Only ${}^{\rm G}M$ particles are re-sampled because the re-sampled particles are used for pose tracking.

%% file: simulation.tex
\section{Simulation experiments}
\label{sec:simulation_experiments}

This section describes experimental results using simulation.
The simulation reffered at Section~\ref{subsubsec:decision_model} was also used for the experiments.

\subsection{Robustness to environment changes}
\label{subsec:robustness_test}

We first tested robustness to environment changes.
The main idea to improve robustness is to introduce the class conditional measurement model denoted as $p({\bf z}_{t} | {\bf x}_{t}, {\bf c}_{t}, {\bf m})$ for likelihood calculation.
Thus, we compared likelihood calculation results in dynamic situations.
We refer the measurement model to CCMM in this section.

In MCL, the likelihood field model~\cite{Thrun_PR} is often used and we refer it to LFM.
LFM is shown in Eq.~(\ref{eq:lfm}).
LFM is cost efficient and enables to calculate smooth likelihood distribution.
However, LFM does not consider large environment changes.
We compare likelihood calculation performance by LFM and CCMM.

\begin{figure}[!t]
    \centering
    \subfloat[Environment in which LFM was used]{\includegraphics[clip, width = 80 mm]{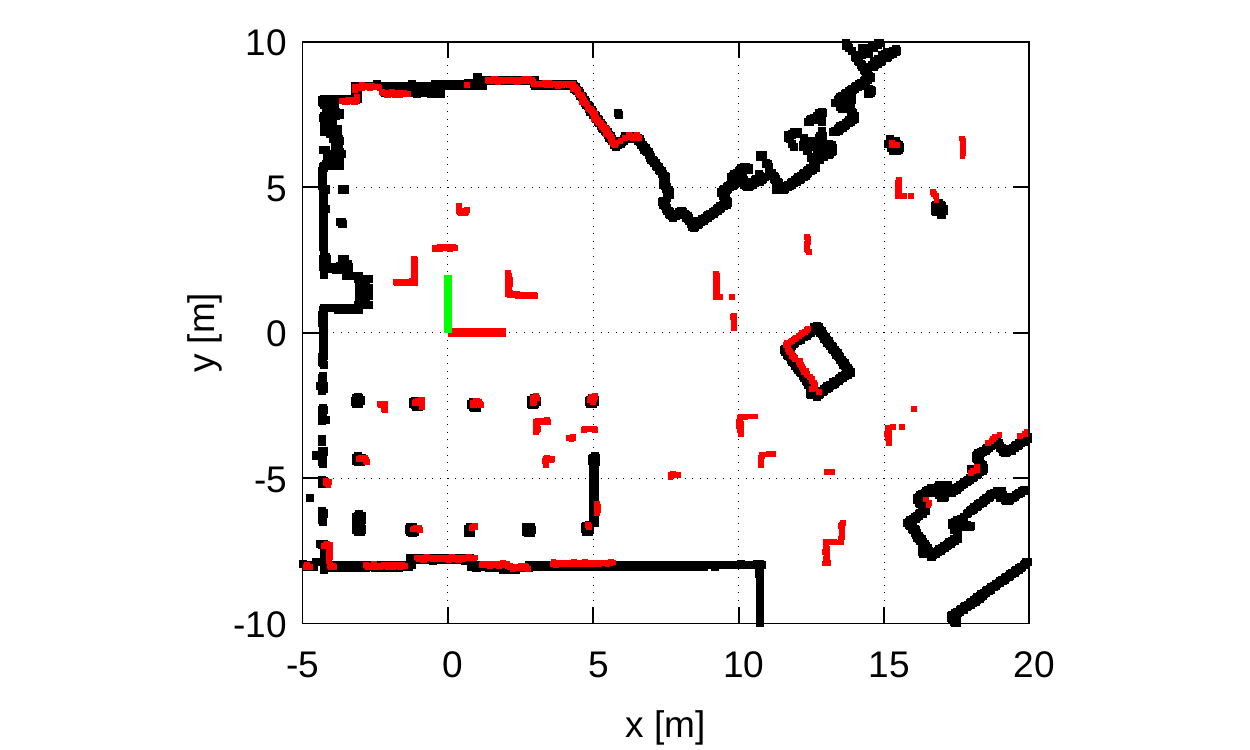}
    \label{fig:lfm_env}}
    \subfloat[Environment in which CCMM was used]{\includegraphics[clip, width = 80 mm]{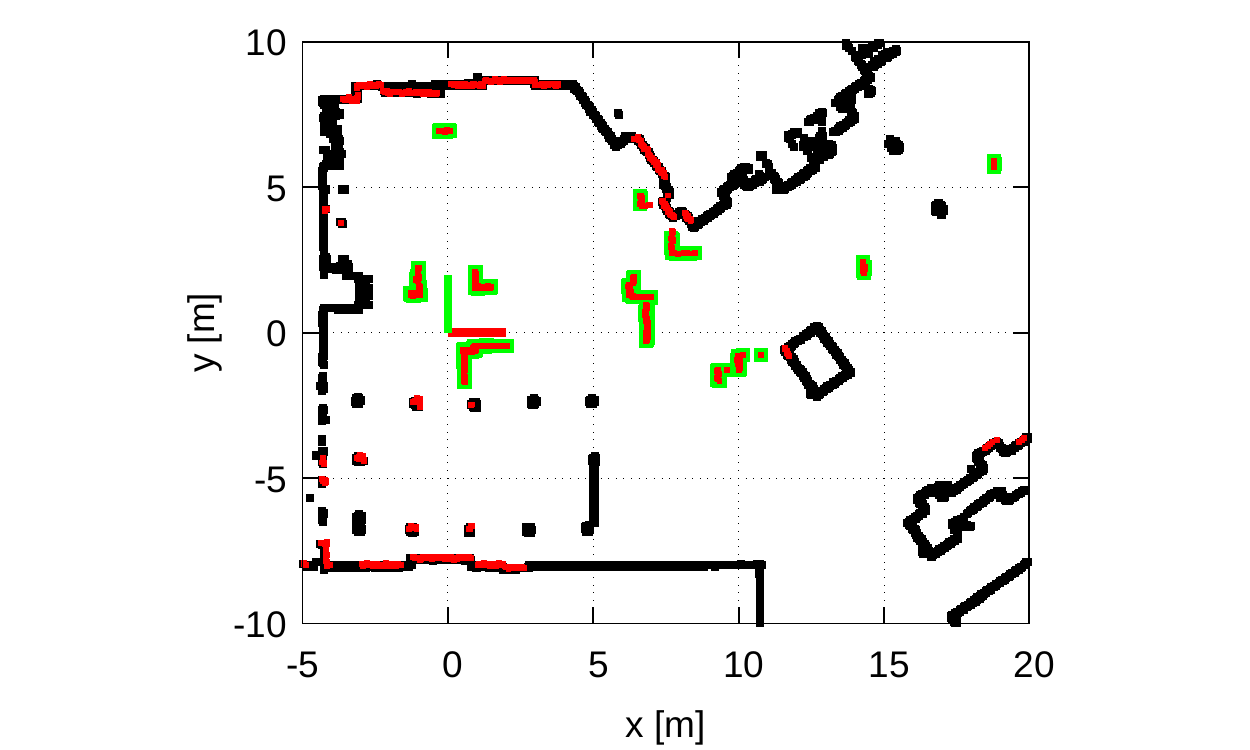}
    \label{fig:ccmm_env}} \\
    \subfloat[A likelihood map by LFM]{\includegraphics[clip, width = 80 mm]{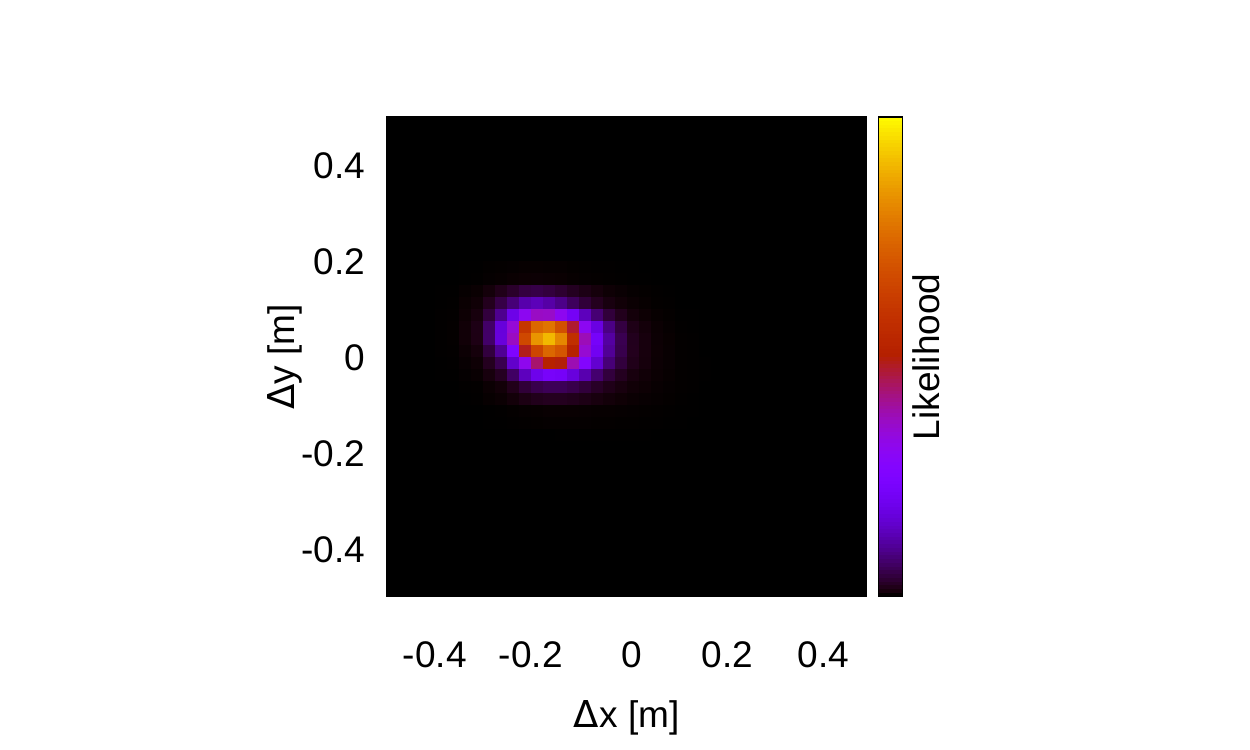}
    \label{fig:lfm_likelihood_map}}
    \subfloat[A likelihood map by CCMM]{\includegraphics[clip, width = 80 mm]{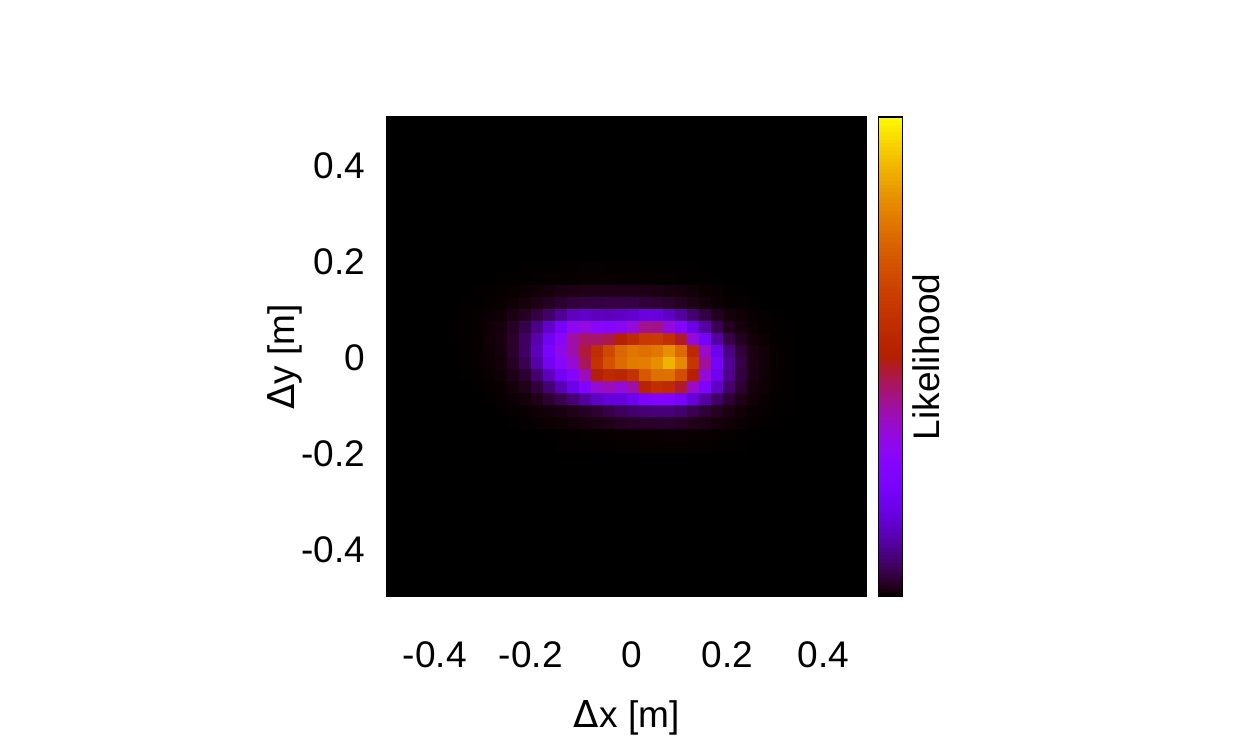}
    \label{fig:ccmm_likelihood_map}} \\
    \caption{Comparison of likelihood maps by LFM and CCMM in dynamic environments. The black, red, and green points of the top figures are the map, scan, and unknown scan points, respectively. The robot locates $(x, y) = (0, 0)$ and faces to $x$-axis parallelly in both the cases. The axes of the likelihood maps represent differences from the ground truth. Note that the heading direction is the same to that of the ground truth.}
    \label{fig:likelihood_map_comparison}
\end{figure}

Figure~\ref{fig:lfm_env} and \ref{fig:ccmm_env} show simulation environments in which LFM and CCMM were used, respectively.
The black and red points indicate map and LiDAR measurement points.
In this simulation, we simulate moving obstacles.
The moving obstacles can be seen from the LiDAR points existing areas in which there are no map points.
Because CCMM also estimates unknown measurements, that is, do not exist on the map, the unknown measurements are illustrated in Fig.~\ref{fig:ccmm_env} with the green.
As can be seen from Fig.~\ref{fig:ccmm_env}, the moving obstacles were classified as the unknown obstacles.

Figure~\ref{fig:lfm_likelihood_map} and \ref{fig:ccmm_likelihood_map} show the likelihood maps around the ground truth.
In both the cases, the ground truth is $(x, y, \theta) = (0, 0, 0)$.
The axes of the likelihood field maps represent differences from the ground truth.
Hence, we can say that likelihood can be robustly calculated if values close to $(\Delta x, \Delta y) = (0, 0)$ are high.
Note that the heading angle is the same to that of the ground truth.

LFM could not generate accurate likelihood distribution; however, CCMM could generate accurate distribution.
In addition, CCMM can be rapidly calculated.
Average likelihood calculation time and its standard deviation were $12.63~{\rm msec}$ and $3.95~{\rm msec}$ when the number of the particles was set to 500.
When LFM is used with the same condition, average and standard deviation are $9.36~{\rm msec}$ and $3.20~{\rm msec}$.
In both the cases, calculation was processed with single thread.
Furthermore, CCMM does not require additional information to calculate LFM, i.e., those memory costs are the same.
From these results, we could confirm that CCMM can increase localization robustness to environment changes without increasing of memory and computational costs.

\subsection{Reliability estimation}

We tested reliability estimation performance.
In the test, we adjusted the noise parameters shown in Eq.~(\ref{eq:motion_model}) to control localization performance.
In other words, localization failures are forcibly occurred owing to misestimate of moving uncertainty.
In addition, the moving obstacles were also simulated as shown in Fig.~\ref{fig:likelihood_map_comparison}.

\begin{figure}[!t]
    \centering
    \subfloat[Estimated trajectory in the success case]{\includegraphics[clip, width = 80 mm]{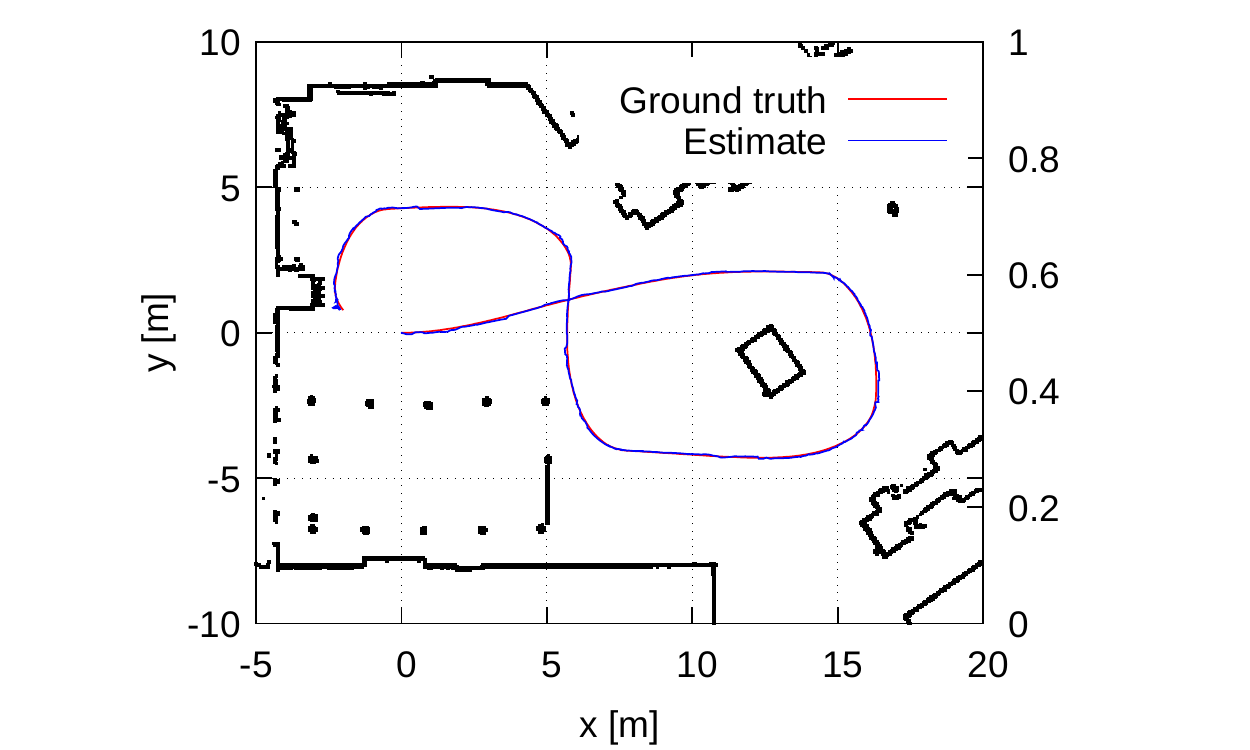}
    \label{fig:success_trajectory}}
    \subfloat[Estimated trajectory in the failure case]{\includegraphics[clip, width = 80 mm]{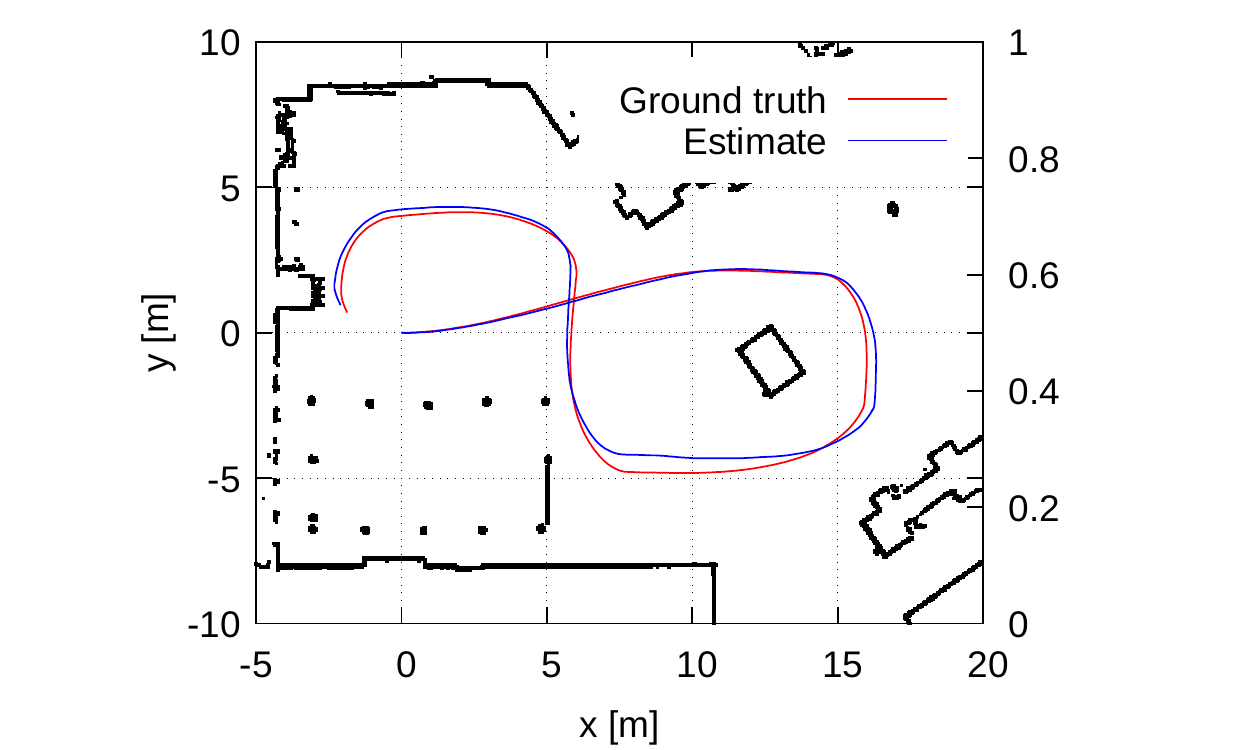}
    \label{fig:failure_trajectory}} \\
    \subfloat[Estimation errors in the success case]{\includegraphics[clip, width = 80 mm]{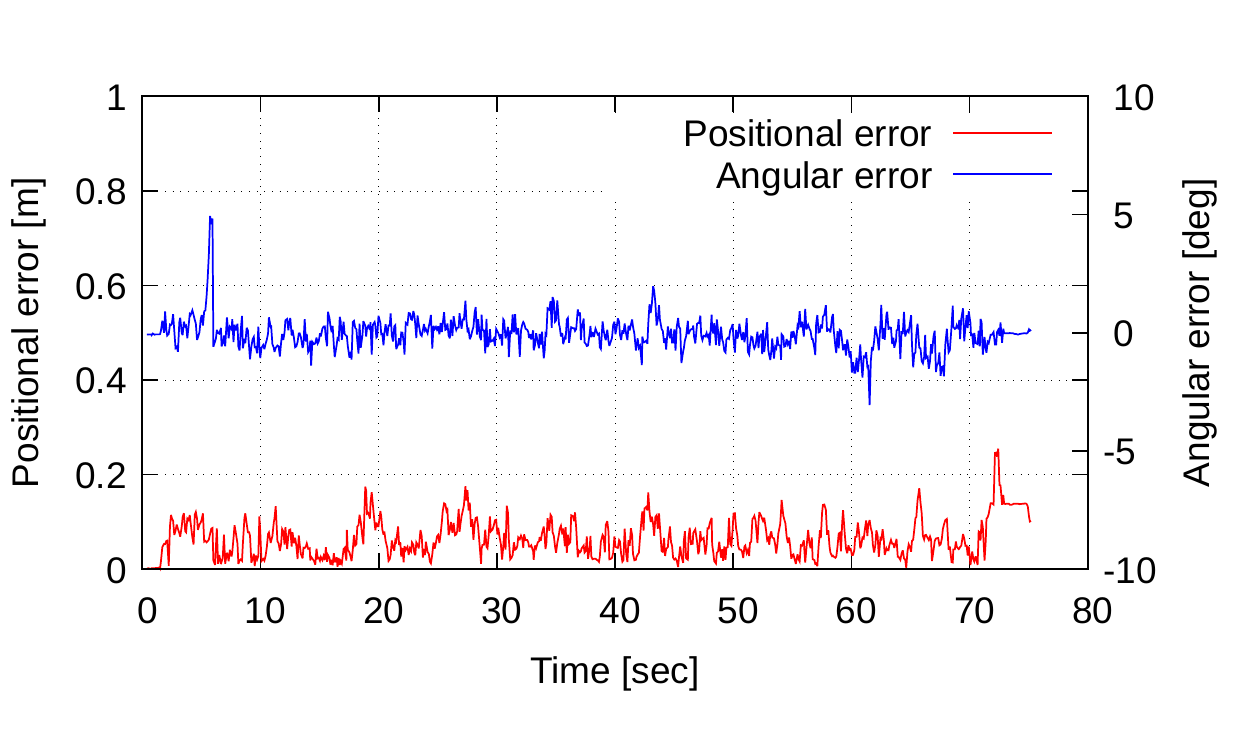}
    \label{fig:success_error}}
    \subfloat[Estimation errors in the failure case]{\includegraphics[clip, width = 80 mm]{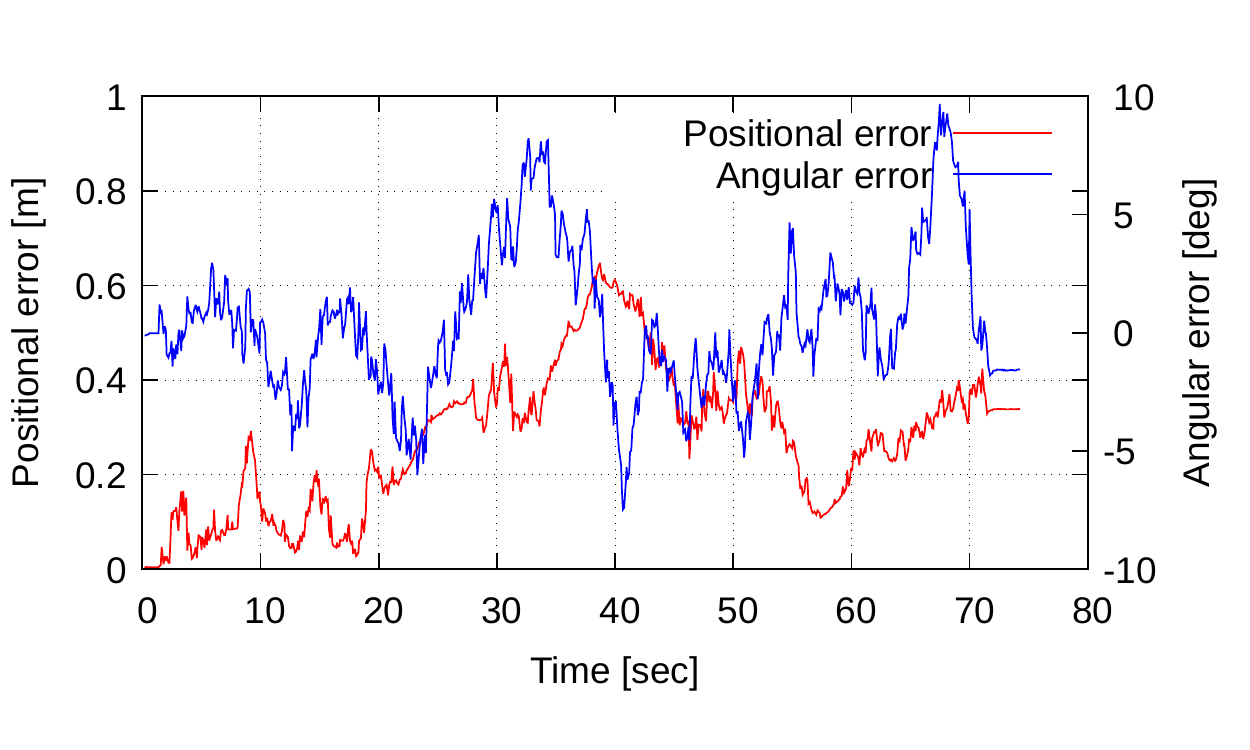}
    \label{fig:failure_error}} \\
    \subfloat[Reliability and MAE in the success case]{\includegraphics[clip, width = 80 mm]{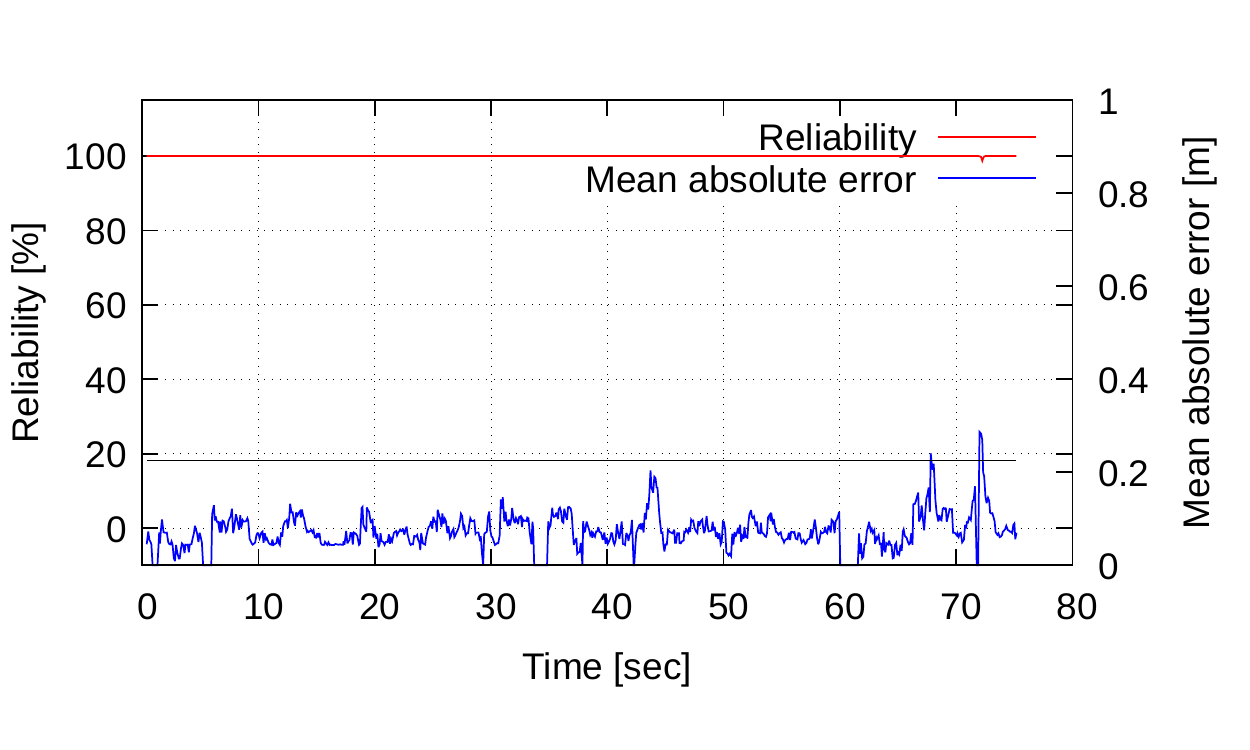}
    \label{fig:success_reliability}}
    \subfloat[Reliability and MAE in the failure case]{\includegraphics[clip, width = 80 mm]{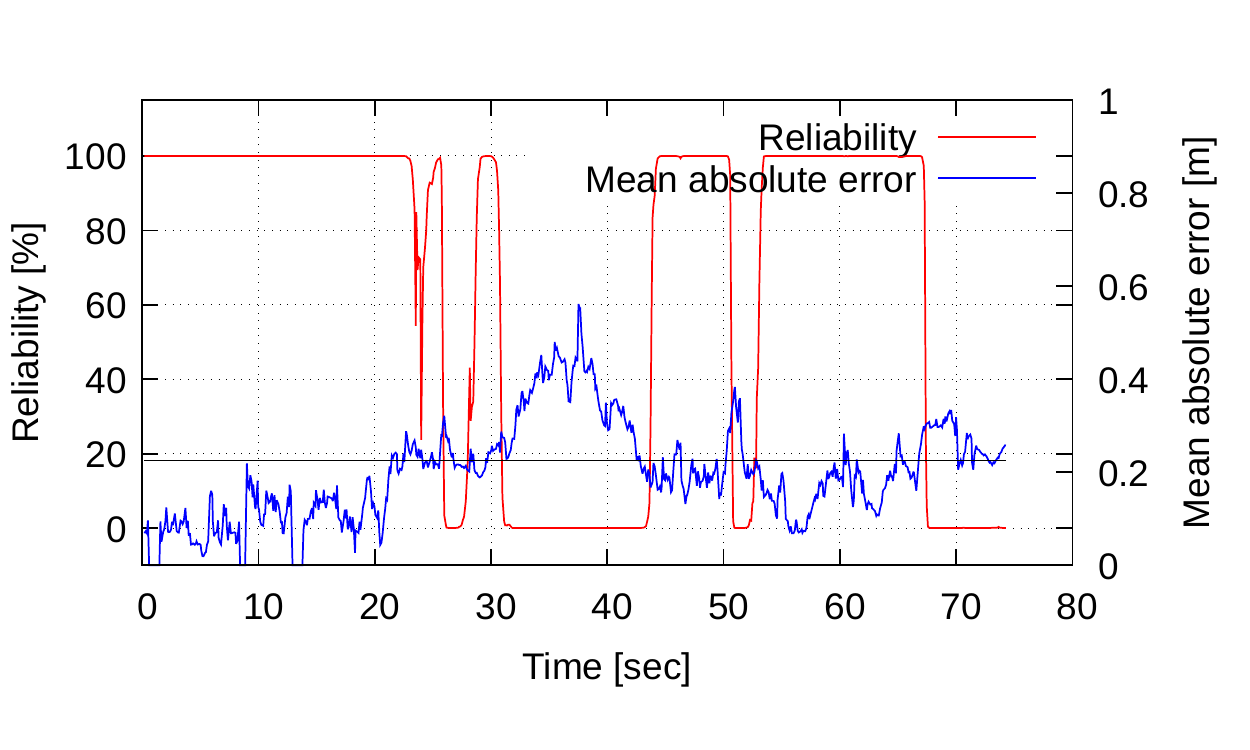}
    \label{fig:failure_reliability}} \\
    \caption{Reliability estimation results in the success (left) and failure (right) cases.}
    \label{fig:reliability_results_simulation}
\end{figure}

The left- and right-hand figures of Fig.~\ref{fig:reliability_results_simulation} show the reliability estimation results in cases where localization has succeeded and failed during the test, i.e., the left- and right-hand figures are the results with the correct and incorrect noise parameters.
In both the cases, the robot followed the same given path.
Figures~\ref{fig:success_trajectory} and \ref{fig:failure_trajectory} indicate the ground truth (red) and estimated (blue) trajectories.
In Fig.~\ref{fig:success_trajectory}, the estimate and the ground truth are the almost same, but these are different in Fig.~\ref{fig:failure_trajectory}.
Figures~\ref{fig:success_error} and \ref{fig:failure_error} show the positional (red) and angular (blue) errors of the estimate from the ground truth in Figs.~\ref{fig:success_trajectory} and \ref{fig:failure_trajectory}.
Figures~\ref{fig:success_reliability} and \ref{fig:failure_reliability} show the estimated reliability (red) and MAE value of the maximum likelihood particle.
The black line shown in Figs.~\ref{fig:success_reliability} and \ref{fig:failure_reliability} indicates the threshold set to MAE.

In the success case shown in the left-hand, the estimation errors were always small and the reliability was always close to $100~\%$.
However, MAE sometimes exceeded the threshold around $70~{\rm sec}$.
Nevertheless, the presented method recognized that localization successfully worked in such a case.
This is because that the presented method estimates the reliability based on Bayes' theorem, i.e., the reliability can be smoothly estimated even when MAE suddenly exceeded the threshold.
The reliability estimation can contribute to improve the classification robustness more than directly using the MAE-based classifier.

In the failure case shown in the right-hand, MAE exceeded the threshold many times.
In particular around $40~{\rm sec}$, MAE completely exceeded the threshold and the presented method estimated that the estimate is unreliable, that is, $0~\%$ reliability.
Hence, the estimated reliability can be used for describing whether localization successfully works or not.

From $20~{\rm sec}$ to $30~{\rm sec}$, the MAE values were less than and exceeded the threshold repeatedly.
In such a case, the neutral reliability value was estimated.
Such a neutral estimation cannot be obtained if the MAE-based classifier is directly used as the classifier.
This is also an advantage to estimate the reliability based on Bayes' theorem.
Based on such neutral estimation, we can consider variety strategy for safety guarantee.

\subsection{Global localization and fusion}

\begin{figure}[!t]
    \begin{center}
        \includegraphics[width = 85 mm]{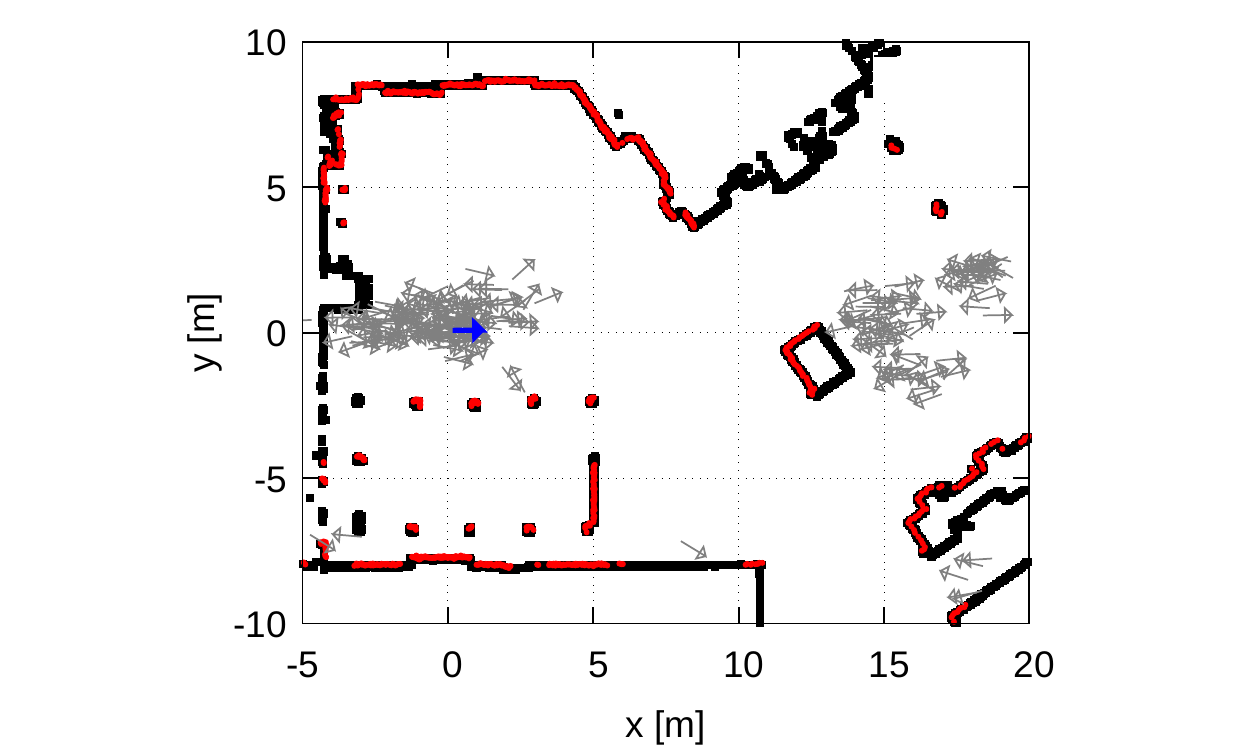}
        \caption{A pose sampling result based on the free-space feature. The black and red points indicate the map and LiDAR measurement points. The blue and gray arrows indicate the ground true and sampled poses.}
        \label{fig:gl_sampling_result}
    \end{center}
\end{figure}

We first tested the pose sampling performance using the free-space feature.
Figure~\ref{fig:gl_sampling_result} shows the pose sampling result.
The blue and gray arrows indicate the ground truth and sampled poses.
The black and red points indicate the map and LiDAR measurement points.
The LiDAR measurements were plotted from the ground truth.
As can be seen from the figure, the pose sampling cannot be accurately performed.
However, some poses were generated around the ground truth.
From the results, we can say that a sophisticated fusion manner is necessary to utilize the pose sampling results.

\begin{figure}[!t]
    \centering
    \subfloat[Before fusion]{\includegraphics[clip, width = 80 mm]{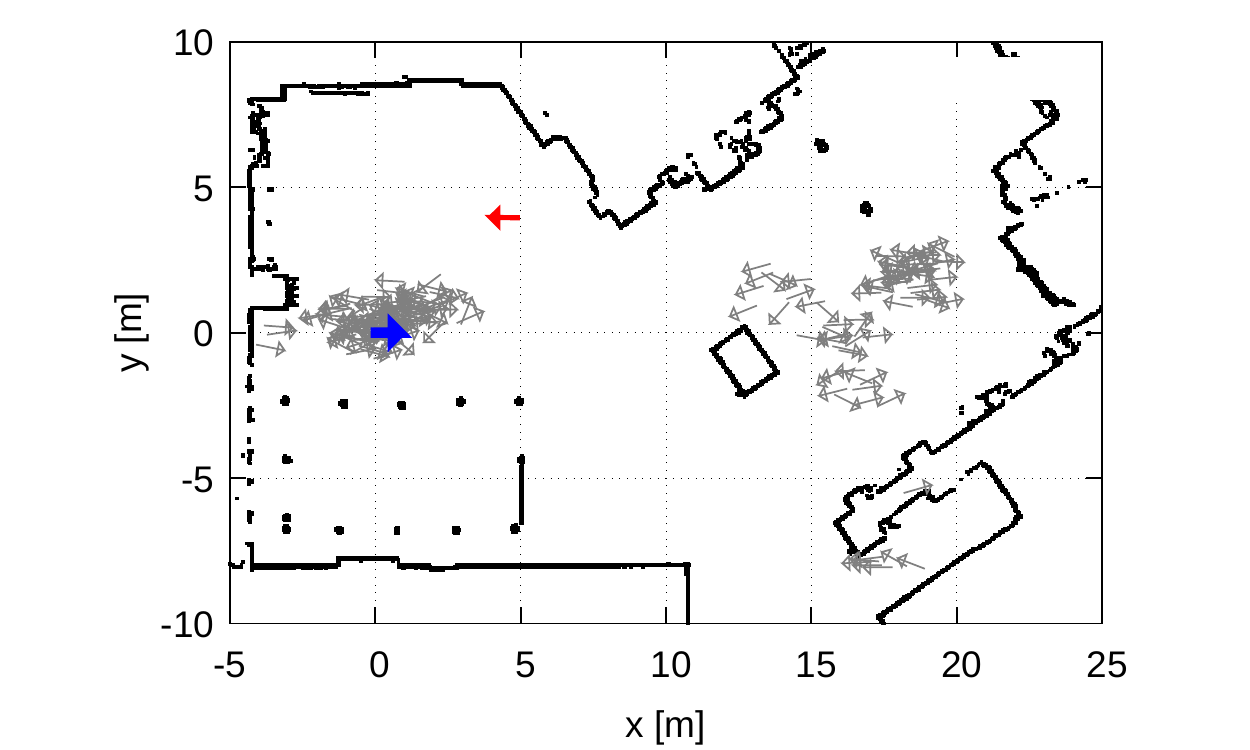}
    \label{fig:gl_sampling_failure_recovery_before}}
    \subfloat[After fusion]{\includegraphics[clip, width = 80 mm]{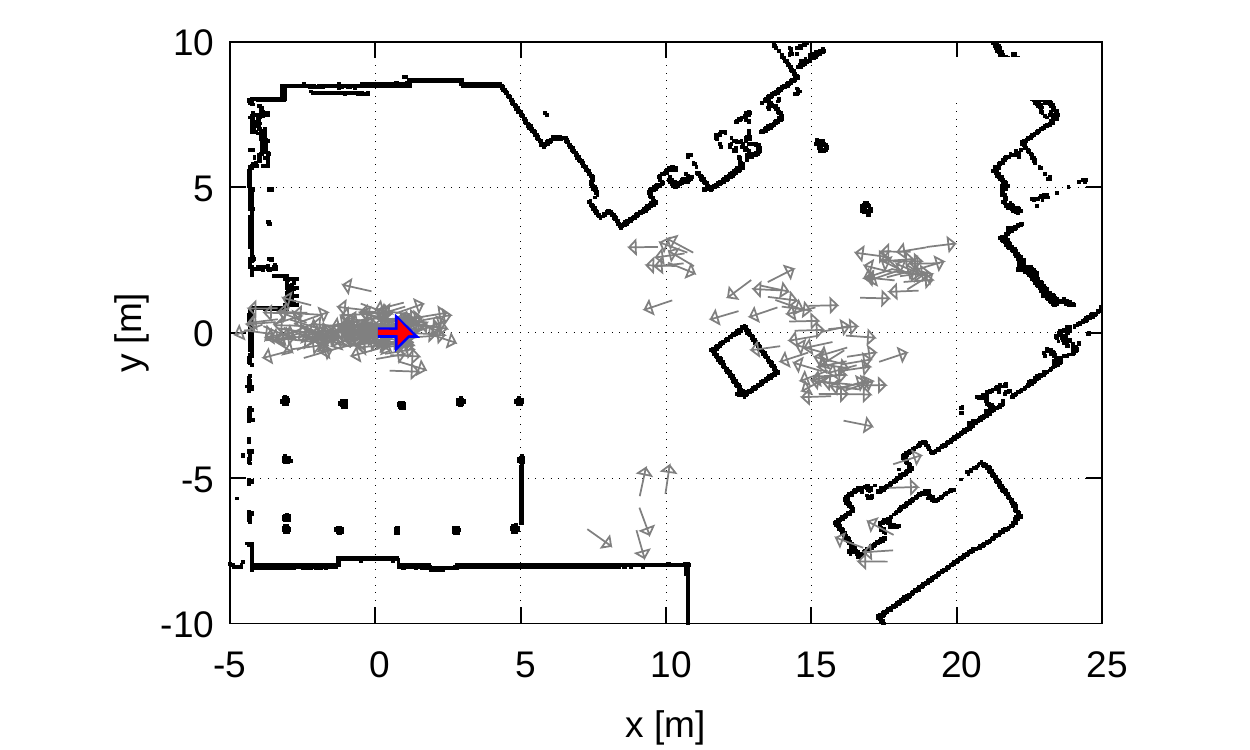}
    \label{fig:gl_sampling_failure_recovery_after}} \\
    \caption{A failure recovery result using the importance-sampling-based fusion. The blue, red, and gray arrows indicate the ground truth, estimate by the presented method, and the sampled poses. The left-hand figure shows the localization failure result, and the right-hand figure shows the case after performing the fusion.}
    \label{fig:gl_sampling_failure_recovery}
\end{figure}

Then, we tested the failure recovery performance using the importance-sampling-based fusion.
Figure~\ref{fig:gl_sampling_failure_recovery} shows the failure recovery result.
The blue, red, and gray arrows indicate the ground truth, estimate by the presented method, and the sampled poses.
Figure~\ref{fig:gl_sampling_failure_recovery_before} shows the localization failure result, and Fig.~\ref{fig:gl_sampling_failure_recovery_after} shows the case after performing the fusion.
As can be seen from the figure, we could confirm that the importance-sampling-based fusion enables to quickly recover from the localization failure.

\begin{figure}[!t]
    \centering
    \subfloat[Estimated trajectory in the success case]{\includegraphics[clip, width = 80 mm]{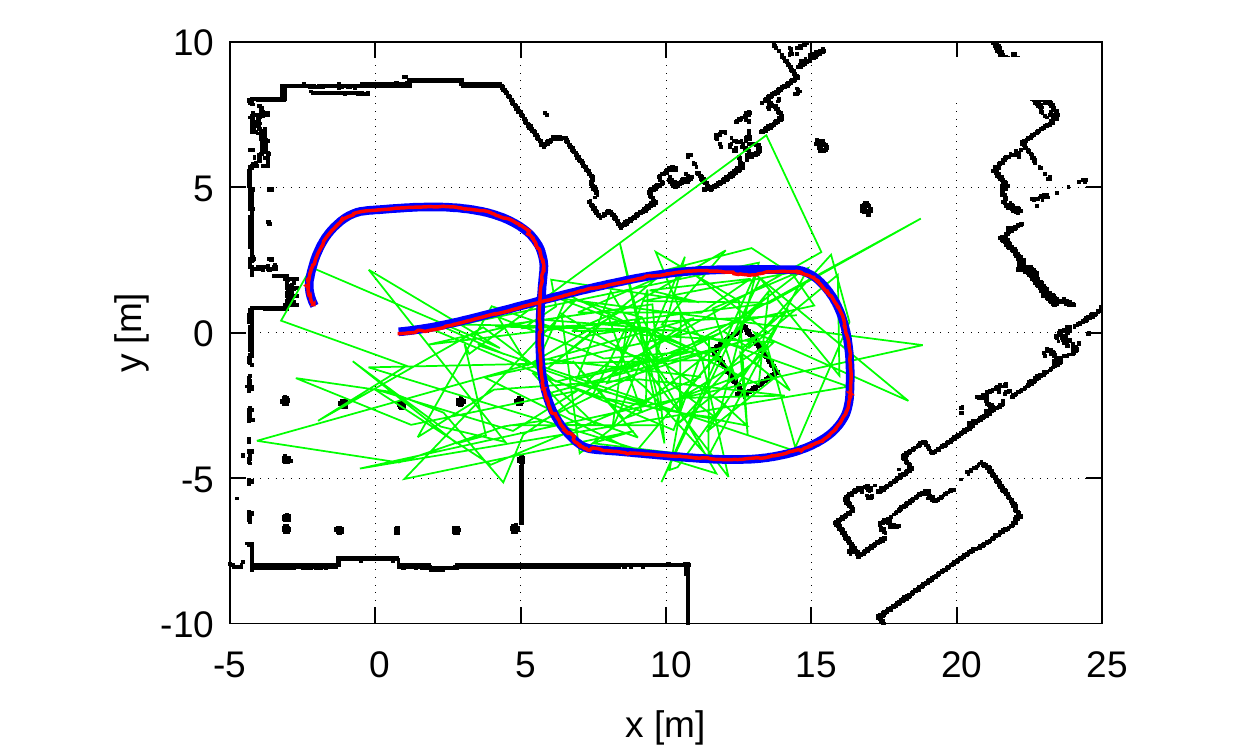}
    \label{fig:gl_sampling_path_following_success}}
    \subfloat[Estimated trajectory in the failure case]{\includegraphics[clip, width = 80 mm]{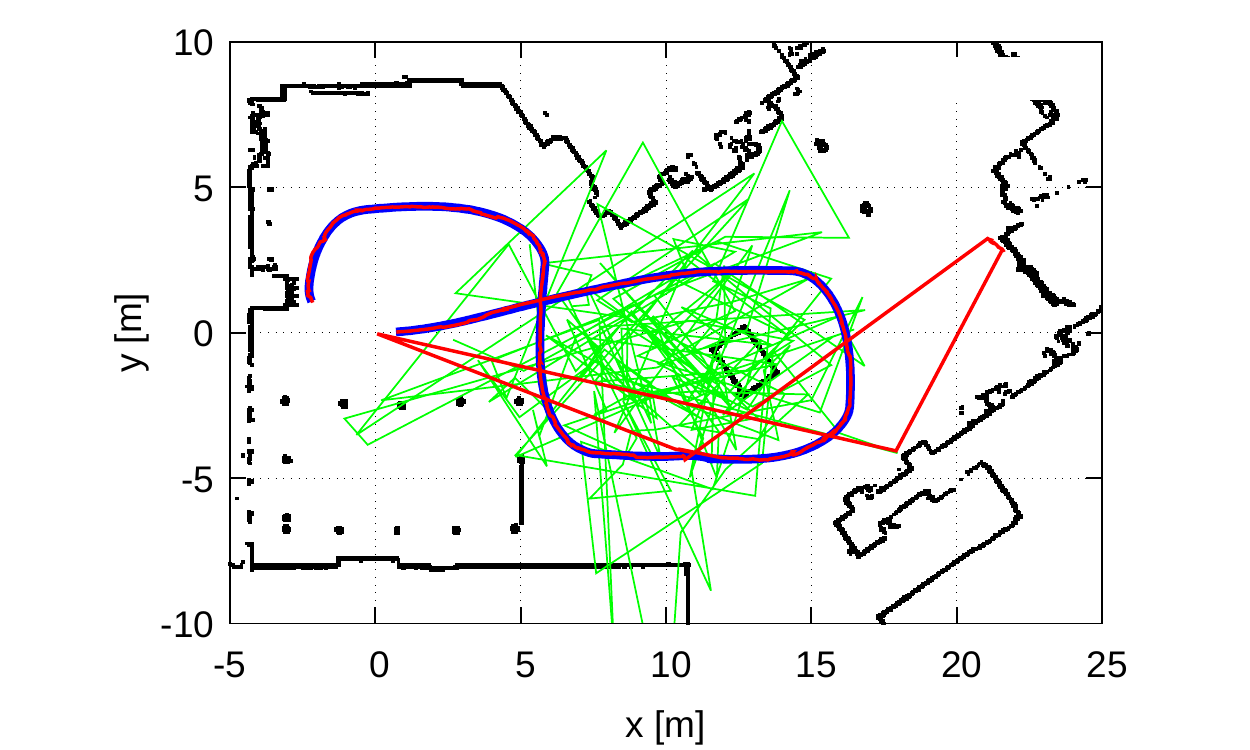}
    \label{fig:gl_sampling_path_following_failure}} \\
    \caption{Pose fusion results with (left) and without (right) the importance sampling. The blue, red, and green lines indicate the ground truth, MCL estimate, and the average of the pose sampling results. The estimate of global localization is the average of the sampled poses.}
    \label{fig:gl_sampling_path_following_results}
\end{figure}

Finally, we tested the fusion performance with and without the importance sampling while the robot moving.
In the cases, the robot also followed the same given path that was used in the tests shown in Fig.~\ref{fig:reliability_results_simulation}.
Figure~\ref{fig:gl_sampling_path_following_results} shows the comparison results of the fusion performance.
The blue, red, and green lines indicate the ground truth, MCL estimate, and the estimate using the free-space-feature-based global localization.
It should be noted that the estimate of global localization is the average of the sampled poses.
In this comparison, we also simulated dynamic obstacles as shown in Fig.~\ref{fig:likelihood_map_comparison}.

In the result shown in Fig.~\ref{fig:gl_sampling_path_following_success}, the importance sampling was used.
The pose sampling could not be accurately performed as shown in Fig.~\ref{fig:gl_sampling_result}; however, the estimate by the presented method could smoothly follow the ground truth.
In the result shown in Fig.~\ref{fig:gl_sampling_path_following_failure}, the importance sampling was not used.
If the importance sampling was not used, the estimates sometimes jumped.
In the likelihood calculation, the measurement-based likelihood calculation, i.e., the class conditional measurement model in the presented method, has the dominant role.
Consequently, pose tracking can work even if the importance sampling is not used.
However, the global-localization-based pose sampling generates poses which make the better match of the LiDAR measurements and map.
In such a case, sudden jump cannot be prevented if the importance sampling is not used.
However, the presented method could prevent such jumps.
From the comparison, we could confirm that the use of the importance sampling contributes to mitigate influence of the noisy estimate by the free-space-feature-based pose sampling.

%% file: experiments.tex
\section{Experiments}
\label{sec:experiments}

In this section, we describe experimental results using public available dataset and our own platform.
We used amcl\footnote{\url{http://wiki.ros.org/amcl}} that is a popular localization package in Robot Operating System (ROS) as a comparison method in all the experiments conducted in this section.
Videos of these experiments can be seen at \url{https://www.youtube.com/watch?v=wsoXvUgJvWk}.

\subsection{Experiments using dataset}

We used Pre-2014 Robotics 2D-Laser Datasets\footnote{\url{https://www.ipb.uni-bonn.de/datasets/}} for the dataset-based experiments.
We selected two datasets; MIT CSAIL and Intel Research Lab.
We first built environment maps using gmaping\footnote{\url{http://wiki.ros.org/gmapping}} and tested the localization methods with wrong initial pose.
Note that we used this software\footnote{\url{https://github.com/NaokiAkai/carmen_player_ros}} to use the datasets in ROS.

We used amcl as a comparison method.
A function used in augmented MCL~\cite{gutmann_iros2002:_amcl} is implemented in amcl.
This function enables to detect localization failure by observing history of likelihoods.
In addition, this function enables to perform re-localization by randomly sampling the particles on the free space if localization failure is detected.
Hence, we compared pose tracking and re-localization performances of the presented method with that of amcl.
However, amcl does not have an estimation function of localization correctness.
We will show the reliability estimation performance by the presented method to reveal its effectiveness.

\begin{figure}[!t]
    \centering
    \subfloat[Ours]{\includegraphics[clip, width = 50 mm]{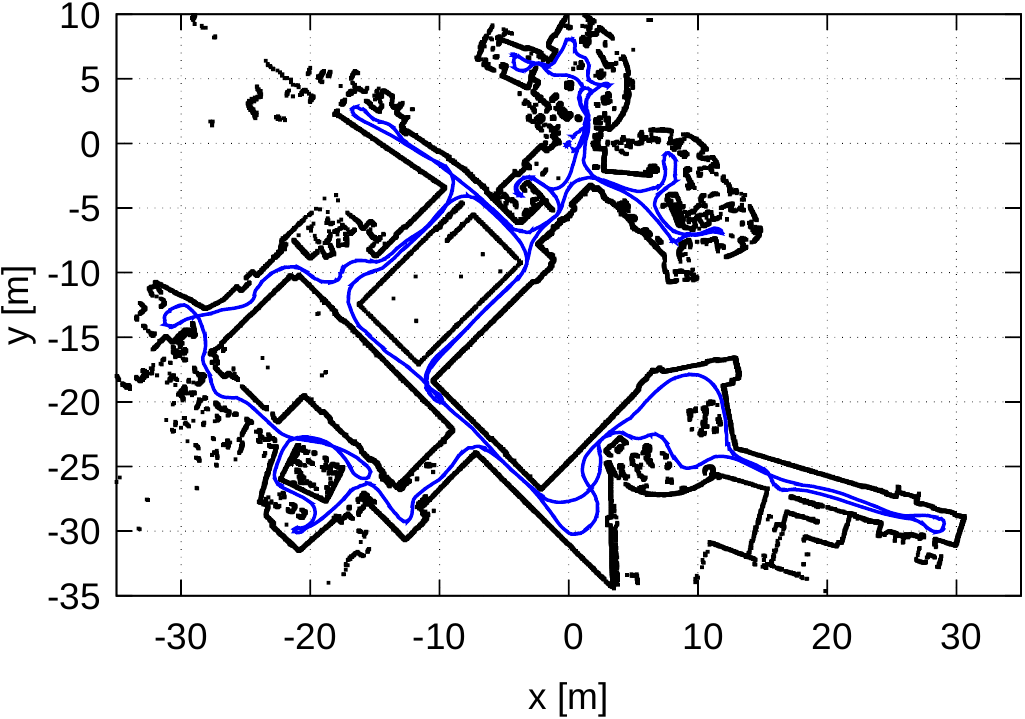}
    \label{fig:als_ros_csail}}
    \subfloat[amcl]{\includegraphics[clip, width = 50 mm]{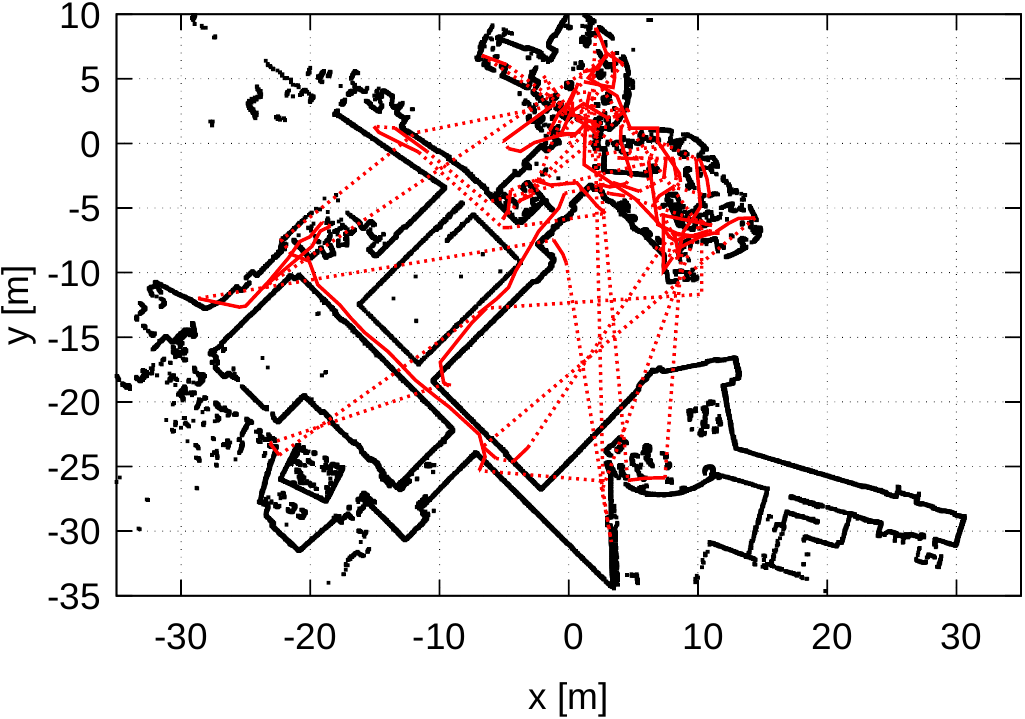}
    \label{fig:acml_csail}}
    \subfloat[Reference]{\includegraphics[clip, width = 50 mm]{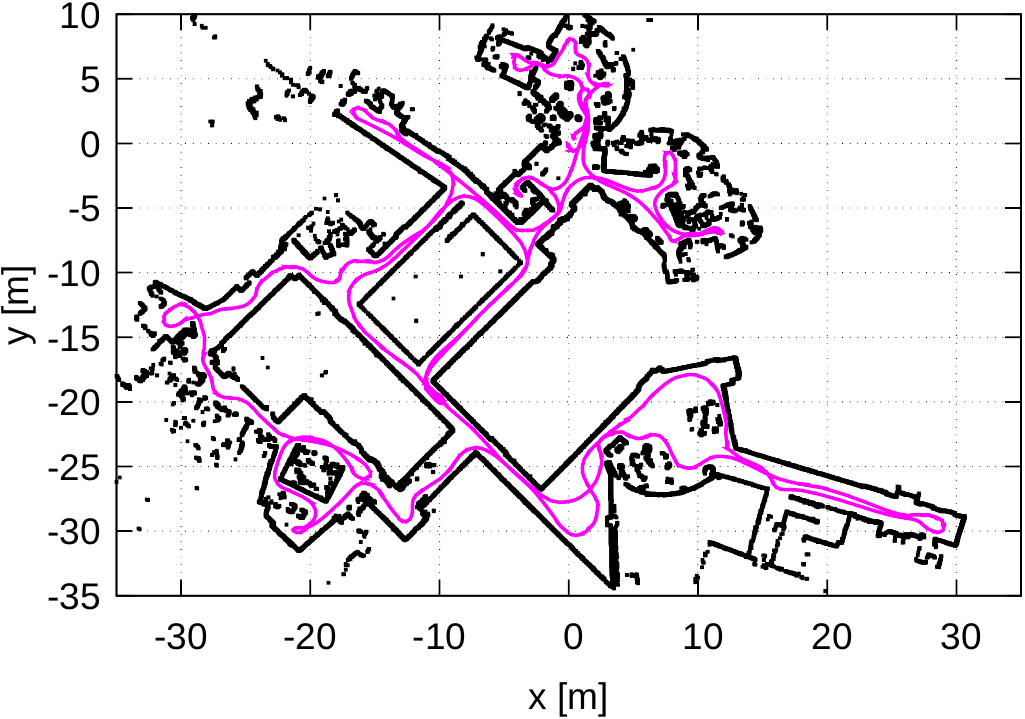}
    \label{fig:reference_csail}} \\
    \caption{The localization results on the MIT CSAIL dataset.}
    \label{fig:localization_experiments_csail}
\end{figure}

\begin{figure}[!t]
    \centering
    \subfloat[Ours]{\includegraphics[clip, width = 50 mm]{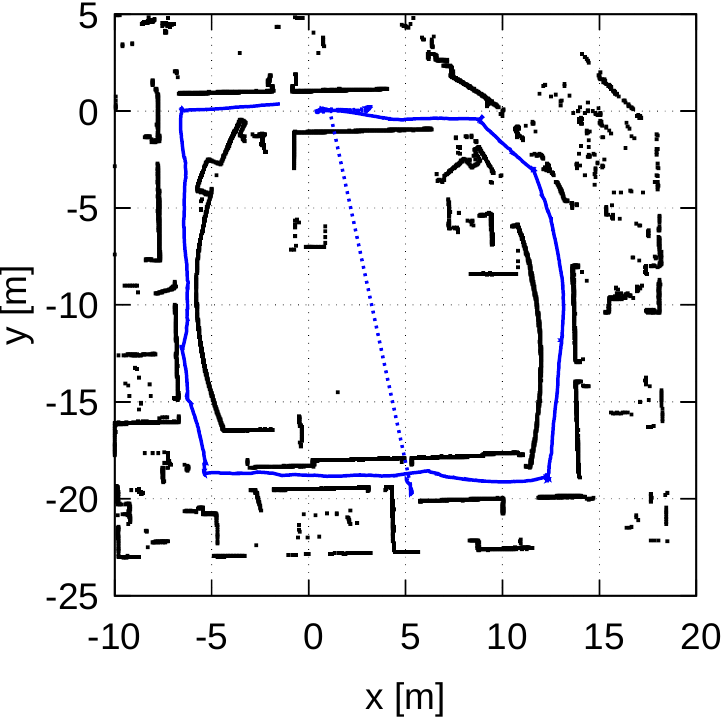}
    \label{fig:als_ros_intel}}
    \subfloat[amcl]{\includegraphics[clip, width = 50 mm]{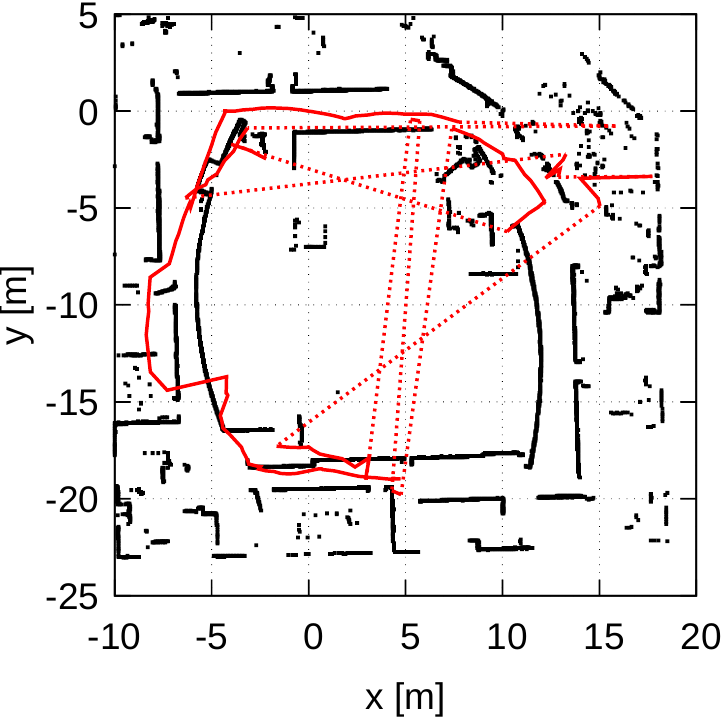}
    \label{fig:acml_intel}}
    \subfloat[Reference]{\includegraphics[clip, width = 50 mm]{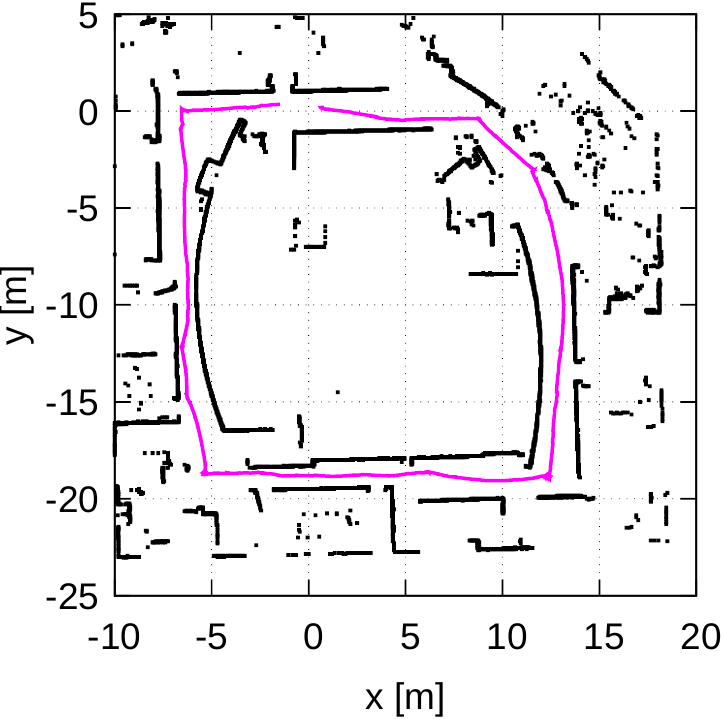}
    \label{fig:reference_intel}} \\
    \caption{The localization results on the Intel Research Lab dataset.}
    \label{fig:localization_experiments_intel}
\end{figure}

Figures~\ref{fig:localization_experiments_csail} and \ref{fig:localization_experiments_intel} show the experimental results on the MIT CSAIL and Intel Research Lab datasets, respectively.
The left and middle figures are the trajectories estimated by the presented method and amcl.
We refer the presented method to ``Ours'' in the captions of this section for notational convenience.
When the localization results are jumped, that is, re-localization was performed, the trajectories are depicted with the dashed lines.
The figures also show the reference trajectories that are estimated by the presented method with the correct initial pose in the right-hand.
We confirmed that the presented method successfully tracked the robot pose while making the reference trajectories. 

In the MIT CSAIL case, the initial position was correct, but the initial heading angle was different about 120~degrees.
The presented method could immediately compensated the heading angle; however, amcl could not compensated the initial error quickly.
After $340~{\rm seconds}$ when the presented method re-localized, amcl could re-localize.
In addition, the presented method achieved accurate pose tracking even though re-localization process was parallelly performed in all the time of the experiment.

In the Intel Research Lab case, the initial pose was completely different.
The correct initial pose was $(x, y, \theta) = (0, 0, 0)$, but the given initial pose was $(x, y, \theta) = (5, -19, \pi)$.
The presented method also quickly re-localized the robot pose; however, amcl could not re-localize the robot pose during the experiment.
In addition, the presented method achieved accurate pose tracking.
From these results, we could confirm that the presented method enables quick re-localization and accurate pose tracking simultaneously.

\begin{figure}[!t]
    \centering
    \subfloat[MIT CSAIL]{\includegraphics[clip, width = 80 mm]{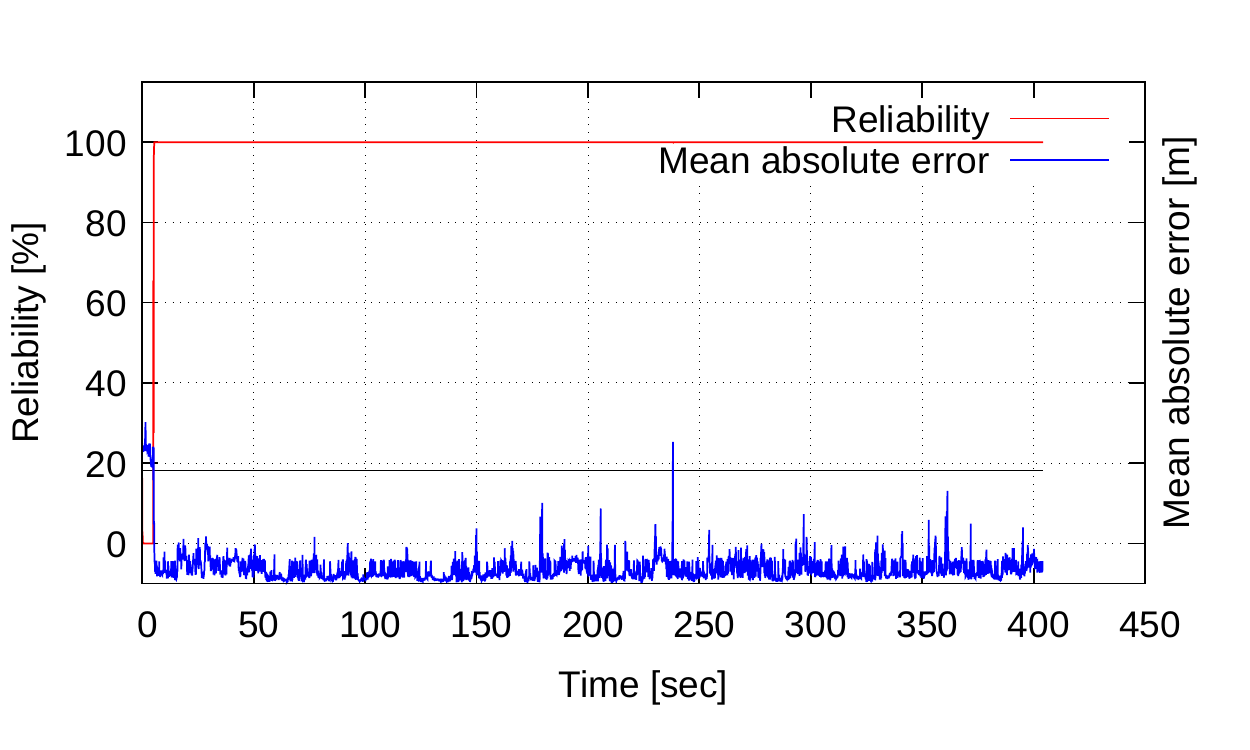}
    \label{fig:reliability_csail}}
    \subfloat[Intel Research Lab]{\includegraphics[clip, width = 80 mm]{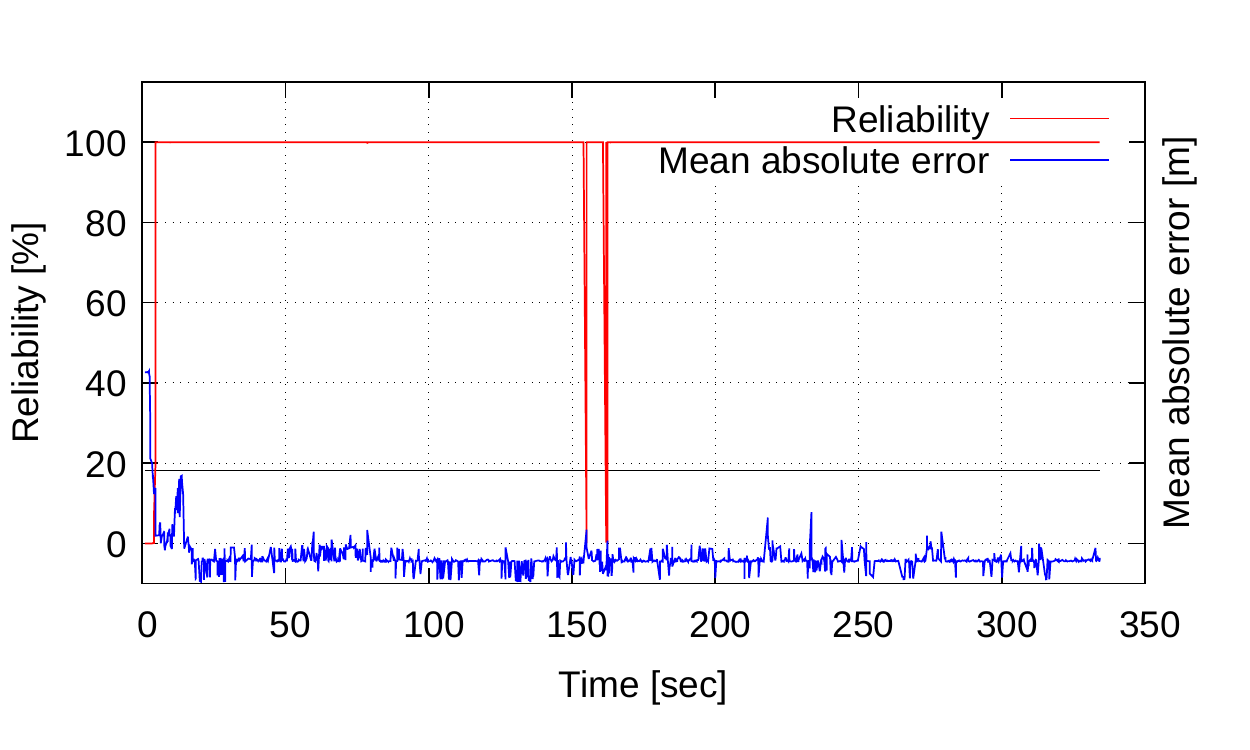}
    \label{fig:reliability_intel}} \\
    \caption{The reliability estimation results on the MIT CSAIL (a) and Intel Research Lab (b) datasets.}
    \label{fig:reliability_results_dataset}
\end{figure}

Furthermore, the reliability estimation results on the MIT CSAIL and Intel Research Lab datasets are shown in Fig.~\ref{fig:reliability_results_dataset}.
In the MIT CSAIL case, low reliability was estimated when around the experiment starts.
This estimation was correct since the wrong initial heading direction was given.
High reliability was estimated after re-localization.
The blue and black lines indicate MAE of the maximum likelihood particle and its threshold.
As can be seen from the figure, MAE exceeded the threshold once; however, reliability was correctly estimated because of reliability was estimated using Bayesian filter.

In the Intel Research Lab case, low reliability was estimated sometimes.
In that cases, the angular error was increased owing to inaccurate INS measurement\footnote{In the Intel Research Lab dataset, translational and angular velocities of the robot are not contained. We calculated the velocities from the pose information and these were sometimes inaccurate owing to significant small differences of timestamps.}.
However, the presented method immediately compensated the estimation error.
From the cases, we can say that localization failures can be immediately detected based on the estimated reliability.
In addition, high reliability was correctly estimated after the error compensation.

\subsection{Experiments with our own platform}

We also conducted experiments with our own experimental platform.
In the experiments, we used Whill Model CR\footnote{\url{https://whill.inc/jp/model-cr}} equipped with UTM-30LX-EW\footnote{\url{https://www.generationrobots.com/en/401435-hokuyo-utm-30lx-ew-laser-range-finder-hokuyo.html}}.
The maximum measurement range of the LiDAR is $30~{\rm m}$ and measurement angle and its resolution are $270~{\rm degrees}$ and $0.25~{\rm degrees}$.
We conducted experiments in an indoor environment with long corridors, i.e., there are areas where enough measurements cannot be obtained to correctly localize the robot pose.
In general, long corridors are difficult areas for performing accurate localization when the maximum measurement range is short.
In the experiments, we also compared the presented method with amcl.

\begin{figure}[!t]
    \centering
    \subfloat[Ours]{\includegraphics[clip, width = 30 mm]{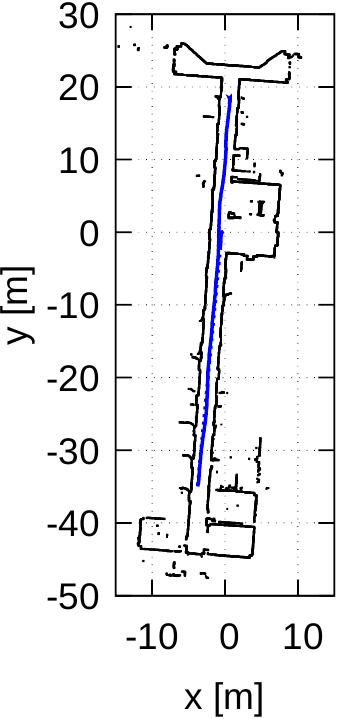}
    \label{fig:als_ros_e2sb4f1}}
    \subfloat[amcl]{\includegraphics[clip, width = 30 mm]{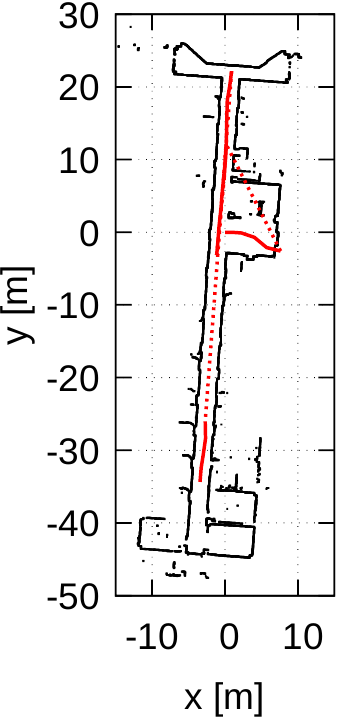}
    \label{fig:acml_e2sb4f1}}
    \subfloat[Reference]{\includegraphics[clip, width = 30 mm]{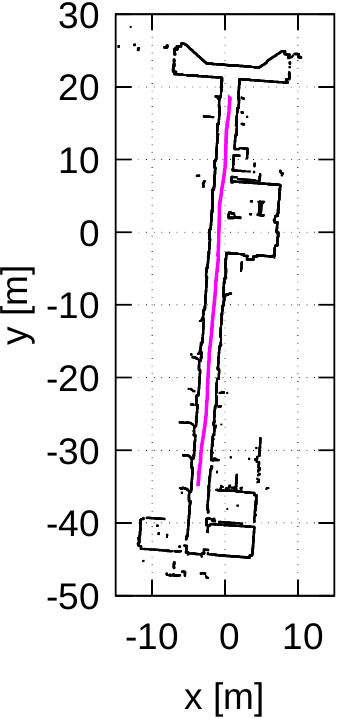}
    \label{fig:reference_e2sb4f1}} \\
    \subfloat[Ours]{\includegraphics[clip, width = 30 mm]{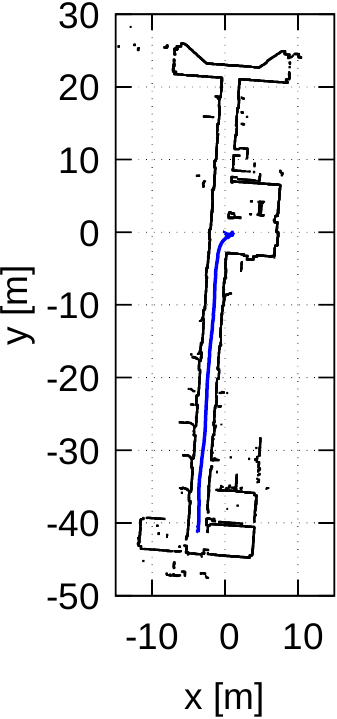}
    \label{fig:als_ros_e2sb4f2}}
    \subfloat[amcl]{\includegraphics[clip, width = 30 mm]{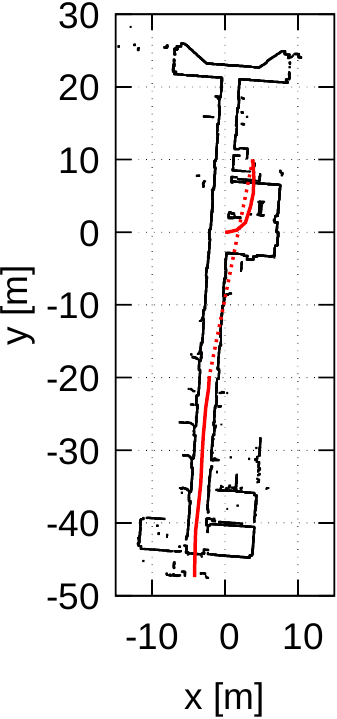}
    \label{fig:acml_e2sb4f2}}
    \subfloat[Reference]{\includegraphics[clip, width = 30 mm]{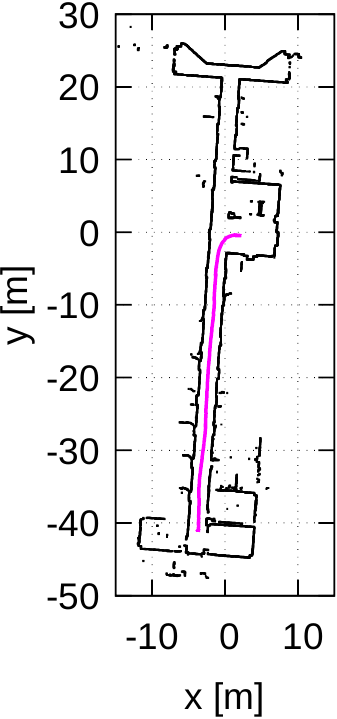}
    \label{fig:reference_e2sb4f2}} \\
    \caption{The localization results in the indoor with two long corridors.}
    \label{fig:localization_experiments_long_corridors}
\end{figure}

Figure~\ref{fig:localization_experiments_long_corridors} shows two experimental results.
In the results, the trajectories estimated by the presented method and amcl, and the reference trajectories are also shown.
In the case shown in Figs.~\ref{fig:als_ros_e2sb4f1} and \ref{fig:acml_e2sb4f1}, the robot started around $(x, y, \theta) = (-4, -35, \pi/2)$; however, the initial pose was set to $(x, y, \theta) = (0, 0, 0)$.
The robot started from the bottom corridor, i.e., effective measurements to compensate the longitudinal error could not be observed.
However, the presented method could re-localize quickly because the rooms shown in the bottom side were observed.
After re-localization was successfully performed, the presented method accurately tracked the robot pose.
However, amcl could not re-localize.
In addition, amcl's estimate sometimes converged to wrong corrdidor side.

In the case shown in Figs.~\ref{fig:als_ros_e2sb4f2} and \ref{fig:acml_e2sb4f2}, the robot started around $(x, y, \theta) = (0, 0, \pi)$; however, the initial pose was set to $(x, y, \theta) = (0, 0, 0)$.
In this case, the presented method also quickly re-localized since the starting area is characteristic in this environment; however, amcl could not re-localize.
From these results, we cloud also confrim that the presented method correctly works in areas with long corridors.

\begin{figure}[!t]
    \centering
    \subfloat[The result in Fig.~\ref{fig:als_ros_e2sb4f1}]{\includegraphics[clip, width = 80 mm]{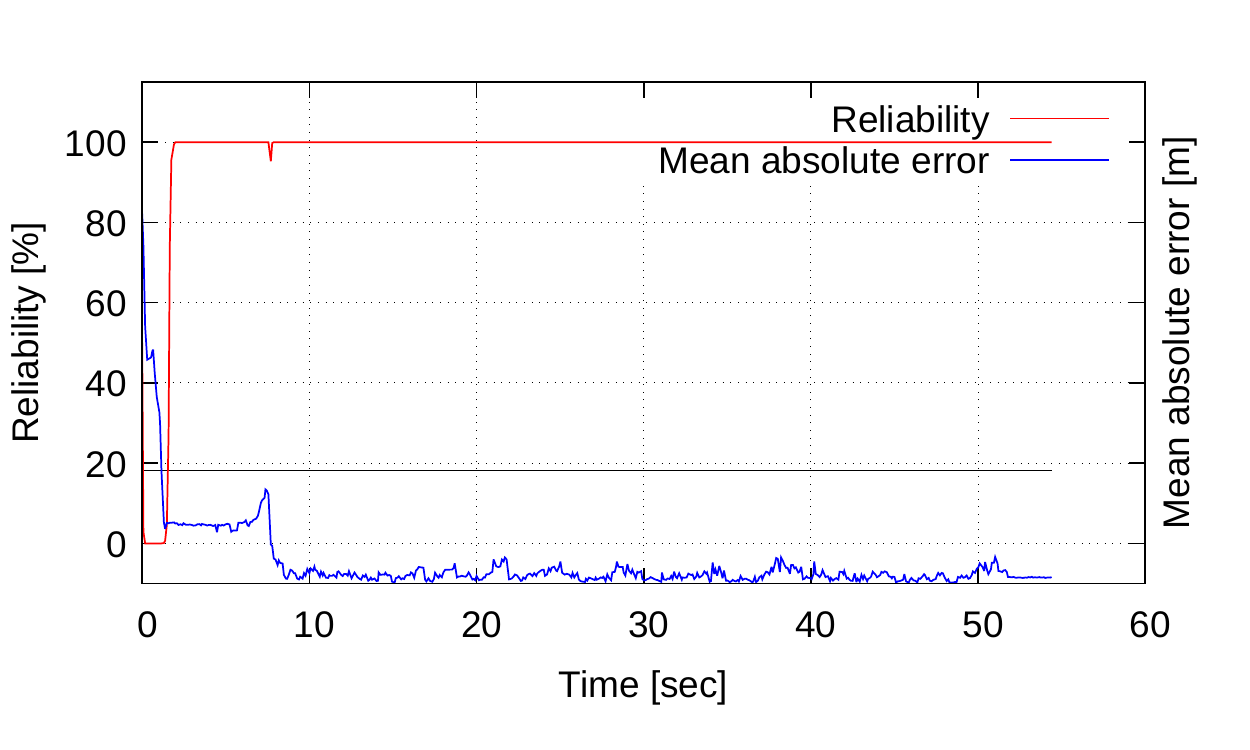}
    \label{fig:reliability_e2sb4f1}}
    \subfloat[The result in Fig.~\ref{fig:als_ros_e2sb4f2}]{\includegraphics[clip, width = 80 mm]{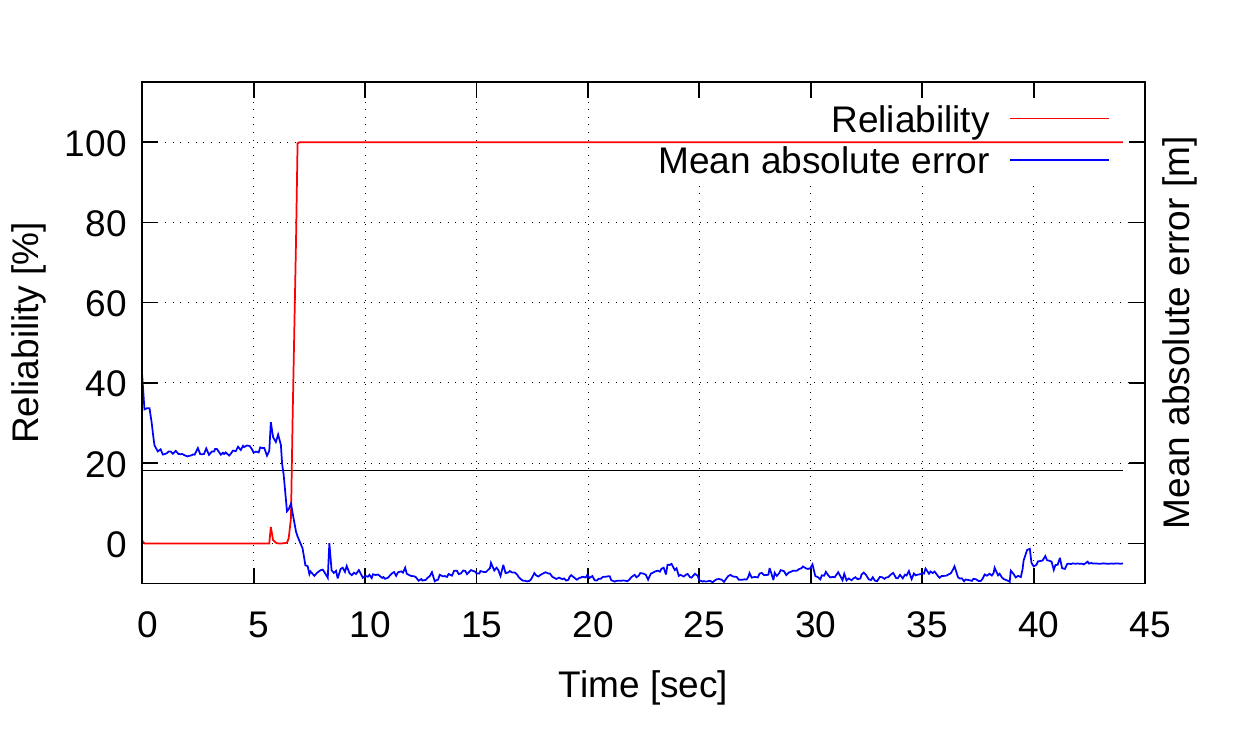}
    \label{fig:reliability_e2sb4f2}} \\
    \caption{The reliability estimation results in the top and bottom cases of Fig.~\ref{fig:localization_experiments_long_corridors}.}
    \label{fig:reliability_results_platform}
\end{figure}

Finally, Fig.~\ref{fig:reliability_results_platform} shows the reliability estimation results in the above two cases.
In both the cases, low reliability was estimated when the experiments start.
These are correct results since the wrong initial pose was given in both the cases.
After re-localization, high reliability was estimated.
These results are also correct.
We confirmed that reliability estimation by the presented method successfully works in the corridor environments.

\subsection{Limitation}

In this subsection, we discuss the limitations of the presented method regarding localization robustness, reliability estimation accuracy, and quickness of re-localization.

\subsubsection{Localization robustness}

To improve localization robustness to environment changes in the presented localization method, we used the class conditional measurement model.
This model has the known and unknown classes as conditional variables.
Owing to the conditional variables, we can separately model the measurements obtained from known and unknown obstacles.
As a result, the measurements can be flexibly modeled and localization robustness is improved without computational cost increasing.
However, this model cannot contribute to realizing guarantee of localization performance since environment changes cannot be perfectly modeled even if the model is used.
Hence, the presented method includes the reliability estimation function.
We strongly note that reliable localization that we consider in this study cannot be realized with only robustness improvement.

The class conditional measurement model has similar effect to M-estimation~\cite{MultipleViewGeometry}.
M-estimation uses a weight function to mitigate influence of outliers.
This weight function gives correspondences having large distance a small weight and their effects are mitigated in the optimization process.
Similar performance to this mitigation is realized owing to the use of the the unknown class in the conditional measurement model.
M-estimation indeed improves robustness; however, it cannot work if initial estimate is far from the ground truth because almost all measurements are considered to be outliers and optimization does not work.
In other words, there is a trade-off relationship between robustness and accuracy in the use of M-estimation.
The class conditional measurement model also yields this negative effect.
Hence, initial guess for localization have to be accurately estimated to utilize the class conditional measurement model.

\subsubsection{Reliability estimation accuracy}
\label{subsubsec:reliability_estimation_accuracy}

In the presented localization method, reliability estimation is performed using the MAE-based classifier.
This classification is performed based on whether the MAE value exceeds the threshold or not.
Of course, this kind of simple-threshold-based classification does not stably work.
However, we achieved stable reliability estimation because the reliability estimation process is performed based on the Bayesian filtering.
Here, stability means ability to cope with noisy classification results.

However, it is hard for the presented method to improve classification accuracy of localization correctness.
Interpretation of reliability estimated by the presented method is that {\it a predicted probability over what localization successfully works in online based on statistical performance of a used classifier}.
Hence, the estimated reliability cannot tell us whether the localization results is close to the ground truth.
The Bayesian filtering indeed contributes to improve stability of estimate because it considers the statistical performance of the classifier.
However, reliability estimation accuracy depends on accuracy of a used classifier.
An accurate classifier is necessary to improve reliability estimation accuracy and simple threshold-based classifier such as MAE-based one might be insufficient in complex scenes.

In~\cite{AkaiRA-L2019}, we presented accurate localization correctness classifier.
However, it is impossible to integrate the classifier with the graphical model shown in Fig.~\ref{fig:graphical_model} because the computational cost of the classifier is high.
To realize these integration, a new graphical model is required.

\subsubsection{Quickness of re-localization}

In this section, we conducted the localization experiments with wrong initial poses.
The presented method achieved quick re-localization from the wrong poses.
However, the re-localization performance depends on pose sampling based on used global localization.
The global localization method used in the presented method utilizes features defined on the free space.
In other words, the global localization performance is affected by environment changes since the features are strongly affected by shape of environment.
We have to improve global localization performance for reliable quick re-localization.

The use of importance sampling improves robustness to miss sampling by global localization.
However, evaluation using the predictive distribution has limitation.
Eq.~(\ref{eq:predictive_distribution_global}) contains the uniform distribution, ${\rm unif}(\cdot)$.
This uniform distribution has effect to achieve re-localization from miss estimation including large error.
However, Eq.~(\ref{eq:predictive_distribution_global}) approximates the predictive distribution, $\int p({\bf x}_{t} | {\bf x}_{t-1}, {\bf u}_{t}) p({\bf x}_{t-1} | {\bf u}_{1:t-1}, {\bf z}_{1:t-1}, d_{1:t-1}, {\bf m}) {\rm d}{\bf x}_{t-1}$, and the use of the uniform distribution means that the robot could exist all the areas even if their movement is predicted using the motion model.
Hence, there is a possibility where the robot jumps even though localization successfully works.
This trade off relationship is limitation of the importance-sampling-based re-localization.
To remove this limitation, accurate global localization is necessary.
However, in general, global localization does not stably work more than pose tracking.
We need to improve the global localization performance to realize accurate and quick re-localization.
Or, we need to estimate exact criterion to use global localization or not such as reliability estimation presented in our proposal.

%% file: conclusion.tex
\section{Conclusion}
\label{sec:conclusion}

This paper has presented reliable MCL that can (1) robustly work in dynamic environments, (2) immediately detect localization failure by estimating reliability, and (3) quickly re-localize if estimate has failed.
To achieve these functions, we presented a novel graphical model and formulate the Bayesian filtering for the simultaneous localization, sensor measurement class estimation, and reliability estimation problem.
We used Rao-Blackwellized particle filter to implement the simultaneous estimation system.

We conducted three types of experiments, that are, simulation-, dataset-, and our-own-platform-based experiments.
Through the simulation experiments, we numerically showed that the presented method can achieve three things mentioned above.
In addition, we showed that the presented method outperforms the traditional approaches.
In the remaining two experiments, we compared the presented method with amcl that is a popular localization package in ROS.
The comparison results showed that the presented method achives fast re-localization more than amcl.
In addition, we confirmed that the presented method can accurately perform pose tracking and reliability estimation.

For the future work, we will try to propose a new graphical model that enables to integrate more accurate localization correctness classifier into the reliability estimation model in order to realize a practical reliability estimator.